\documentclass[10pt,twocolumn,letterpaper]{article}

\usepackage[final,applications]{wacv}

\usepackage[utf8]{inputenc}
\usepackage[T1]{fontenc}
\usepackage{threeparttable}
\usepackage{tabularx}
\usepackage{longtable}
\usepackage{array}
\usepackage{siunitx}
\usepackage{float}
\usepackage{placeins}

\graphicspath{{figures/}}

\definecolor{wacvblue}{rgb}{0.21,0.49,0.74}
\usepackage[pagebackref,breaklinks,colorlinks,allcolors=wacvblue]{hyperref}

\def\wacvPaperID{2062}
\def\confName{WACV}
\def\confYear{2027}
\usepackage{orcidlink}

\usepackage{etoolbox}
\makeatletter
\patchcmd{\@maketitle}{\@author}{\@author\\[0.25em]\@affiliation}{}{}
\makeatother

\title{Integrating Unimodal and Vision--Language Representations in Latent Space for Multi-Label Chest X-Ray Classification}

\author{%
  Quang-Huy Tran$^{1,2,3}$~\orcidlink{0009-0005-3172-9093},
  Duc-Tuan Ngo$^{1,2,3}$~\orcidlink{0009-0005-6854-8632},
  Minh-Khoi Nguyen-Bui$^{1,2,3}$\orcidlink{0009-0002-3419-8097},
  Dang-Khoa Bui$^{1,2,3}$~\orcidlink{0009-0007-7839-2437},\\
  Thanh-Trong Tran$^{1,2,3}$\orcidlink{0009-0004-6962-4425},
  Tuan-Khoi Nguyen$^{1,2}$~\orcidlink{0000-0001-8556-5876},
  Hoang-Anh Ngo$^{3}$~\orcidlink{0000-0002-7583-753X}
}

\affiliation{%
  $^{1}$Faculty of Computer Science and Engineering, Ho Chi Minh City University of Technology, \\ 268 Ly Thuong Kiet Street, Dien Hong Ward, Ho Chi Minh City, Vietnam\\
  $^{2}$Vietnam National University Ho Chi Minh City,
Dong Hoa Ward, Ho Chi Minh City, Vietnam\\
  $^{3}$ Applied AI \& Data Science Division, AK Technologies Company Limited, Ho Chi Minh City, Vietnam \\[0.05cm]
  \ttfamily{ \small \{huy.tranquanghcmutk22, tuan.ngotuan, khoi.nguyenkwan0103, khoa.buidkhoa} \\
  \ttfamily{ \small trong.tran588413799, tuankhoin\}@hcmut.edu.vn, hoang-anh.ngo@aktech.ai.vn}%
}

\begin{document}
\maketitle

\begin{abstract}
Multi-label chest X-ray classification has attracted considerable attention in recent years, with the effective use of visual representations and clinical semantic knowledge playing an important role. This study proposes a framework that combines unimodal representations from RAD-DINO with vision--language representations from BioViL-T for the classification of 14 labels in the MIMIC-CXR-JPG dataset. The RAD-DINO and BioViL-T embeddings and their combined representation are refined separately in latent space before being normalized and fused across the three branches. In addition to improving classification performance, the study aims to clarify the role of each embedding source and the degree to which they complement one another. 

Experiments show that RAD-DINO outperforms BioViL-T when used independently, whereas early fusion further improves the results, indicating that the two embedding sources contain complementary information. The best-performing model achieves a mean AUROC of \(0.840\) and an mAP of \(0.467\). Ablation analysis shows that hybrid fusion provides consistent and statistically significant improvements over early fusion when each embedding source is refined in latent space, suggesting that fusion effectiveness depends on the quality of the representation supplied by each branch. However, the study has only been evaluated internally on MIMIC-CXR-JPG; its generalizability to data from other healthcare institutions therefore remains to be validated. The source code is available at: \href{https://anonymous.4open.science/r/mimic-report-C210/}{https://anonymous.4open.science/r/mimic-report-C210/}.
\end{abstract}

\noindent\textbf{Keywords:} chest X-ray, multi-label classification, RAD-DINO, BioViL-T, latent space, MIMIC-CXR-JPG.

\section{Introduction}
\label{sec:introduction}

\subsection{Background and significance}

Chest X-ray (CXR) is widely used for assessing cardiopulmonary disease and is
commonly formulated as a multi-label classification problem because multiple
abnormalities or support devices may appear simultaneously
\cite{broder2011chest,chen2020labelcooccurrence}. Large-scale datasets such as
MIMIC-CXR-JPG, containing 377,110 images from 65,379 patients, have enabled
the development of increasingly capable CXR classifiers
\cite{johnson2019mimicjpg}.

Earlier approaches relied mainly on CNN architectures such as DenseNet and
CheXNet \cite{huang2017densenet,rajpurkar2017chexnet}, whereas recent work
increasingly uses foundation encoders pretrained on biomedical data. RAD-DINO
learns transferable visual representations from image-only pretraining
\cite{perezgarcia2025raddino}, while BioViL-T uses vision--language and
temporal supervision from chest X-rays and radiology reports
\cite{bannur2023biovilt}. These different pretraining objectives motivate
investigating whether their representations provide complementary information
for downstream CXR classification.

\subsection{Research gap and aim}

RAD-DINO may preserve visual information underrepresented in radiology reports,
although its representations are not explicitly aligned with clinical
language \cite{perezgarcia2025raddino}. BioViL-T instead encourages stronger
alignment with radiological semantics through image--report and temporal
associations, but may place less emphasis on unreported visual signals
\cite{bannur2023biovilt}. We therefore hypothesize that the two representation
sources are complementary.

We examine whether combining RAD-DINO and BioViL-T improves over either source
alone, whether separate latent refinement improves hybrid fusion, and how the
three branches contribute to prediction. Our contributions are a systematic
evaluation of this complementarity in an image-only inference setting, a
latent hybrid fusion framework that refines the three representation branches
before prediction fusion, and ablation and label-wise analyses that localize
the resulting gains.

All experiments use frozen pretrained encoders and require only images at
inference time. Evaluation is performed on the 14 CheXpert observation labels
provided with MIMIC-CXR-JPG. The reported DenseNet121 result is used only as
an external reference rather than as a baseline trained under our experimental
protocol.
\section{Related work}

\subsection{CNN and Transformer-Based Multi-Label CXR Classification}
Building on CNN backbones, many studies have sought to improve multi-label CXR
classification by better exploiting spatial information and relationships
among diseases and by addressing class imbalance. These are important
challenges because multiple abnormalities may co-occur in one image while
varying substantially in size, location, and prevalence~\cite{ge2018chestxray}.
Subsequent approaches introduced attention mechanisms, modeled relationships
among labels, or adjusted learning strategies to improve the detection of
less prevalent diseases. For example, recent methods exploit label
correlations, label-specific features, and attention to improve the
representation of each disease~\cite{li2024mbranet, ZHANG2024108032}.

Transformer-based architectures further extend this approach through their
ability to model interactions among image regions over a broader context.
SwinCheX uses a Swin Transformer for multi-label CXR classification, whereas
LT-ViT exploits interactions between image tokens and label-representative
tokens. HydraViT further combines a Transformer with disease-specific branches
to capture both global context and relationships among labels~\cite{10222175,
OZTURK2025106959}. Overall, the progression from CNNs to Transformers has
primarily focused on improving how features from a single visual
representation source are exploited and interact.

\subsection{Foundation Models for CXR Representation}
RAD-DINO is pretrained entirely on unimodal biomedical image data. The model
inherits the DINOv2 architecture with a ViT-B/14 Vision Transformer backbone,
in which an image is divided into patches and encoded as a sequence of tokens.
Global image representations and local features are learned jointly through
self-distillation and masked image modeling, enabling RAD-DINO to support
classification and segmentation tasks with performance comparable to or
better than language-supervised biomedical models
\citep{perezgarcia2025raddino,oquab2024dinov2}. BioViL-T, meanwhile, uses a
CNN--Transformer encoder to extract local features and learn interactions
between current and prior images. Radiology reports are modeled with CXR-BERT,
after which the image and text representations are mapped into a shared space
for multimodal learning. The two branches are jointly trained to produce
vision--language representations that reflect both image content and temporal
progression~\citep{bannur2023biovilt}.

\subsection{Latent-space Representations}
Latent space makes it possible to compress data into a lower-dimensional
representation while retaining important features. Variational autoencoders
are a representative approach, whereas latent diffusion performs diffusion
directly on compressed representations to reduce computational cost
~\citep{kingma2014autoencoding,rombach2022latent}. In the medical domain,
DDL-CXR uses a VAE and diffusion in latent space to create CXR representations
suited to the patient's state at prediction time while encouraging the latent
representation to retain information related to clinical abnormalities
~\citep{yao2024ddlcxr}. Another related approach is the Perceiver, which uses a
latent bottleneck to map the input to a smaller set of representations before
further attention-based processing~\citep{jaegle2021perceiver}. These studies
show that latent space can be used not only to compress data but also to refine
and reorganize representations before downstream tasks.

Overall, prior work has separately examined many relevant components, ranging
from CNN- and Transformer-based CXR classification and representation learning
with foundation encoders to latent-space representation refinement. However,
combining purely visual RAD-DINO embeddings with BioViL-T vision--language
embeddings, refining each source in latent space, and performing hybrid fusion for multi-label CXR classification remain underexplored. This gap
is the focus of the present study.


\section{Methodology}
\label{sec:methodology}
\subsection{Research design}
\label{subsec:research_design}

Five configurations are compared:

\begin{enumerate}[label=(\roman*)]
    \item A 768-dimensional \textbf{RAD-DINO only}.

    \item A 128-dimensional \textbf{BioViL-T only}.

    \item \textbf{Early fusion (CONCAT):}
    direct concatenation of the RAD-DINO and BioViL-T embeddings before classification.

    \item \textbf{Hybrid fusion (no latent refinement):}
    normalized logits from the RAD-DINO, BioViL-T, and CONCAT branches are fused directly.

    \item \textbf{Latent hybrid fusion:}
    the three branches are refined separately in latent space before their normalized logits are fused.
\end{enumerate}

\subsection{Data sources and study cohort}

The MIMIC-CXR-JPG data were de-identified and accessed in accordance with
PhysioNet credentialing requirements~\cite{PhysioNet-mimic-cxr-2.1.0}. This
study uses secondary data and involves no direct interaction with patients.
After filtering, the final cohort contains 211,130 distinct studies, each
processed according to the multi-view rule described in
Section~\ref{subsec:multi_view_processing}. Each study is then represented by
two feature vectors corresponding to RAD-DINO and BioViL-T. The early-fusion
representation is constructed by concatenating these vectors. Each embedding
type is independently standardized with \texttt{StandardScaler}, whose
statistics are estimated exclusively from the training set. The RAD-DINO and
BioViL-T backbones remain frozen throughout all experiments; only the
subsequent processing and classification layers are trained in each
configuration.

\subsection{Latent hybrid fusion architecture}
\label{subsec:latent_label_wise_fusion}

\begin{figure*}[t]
    \centering
    \includegraphics[
        width=0.75\textwidth,
        keepaspectratio
    ]{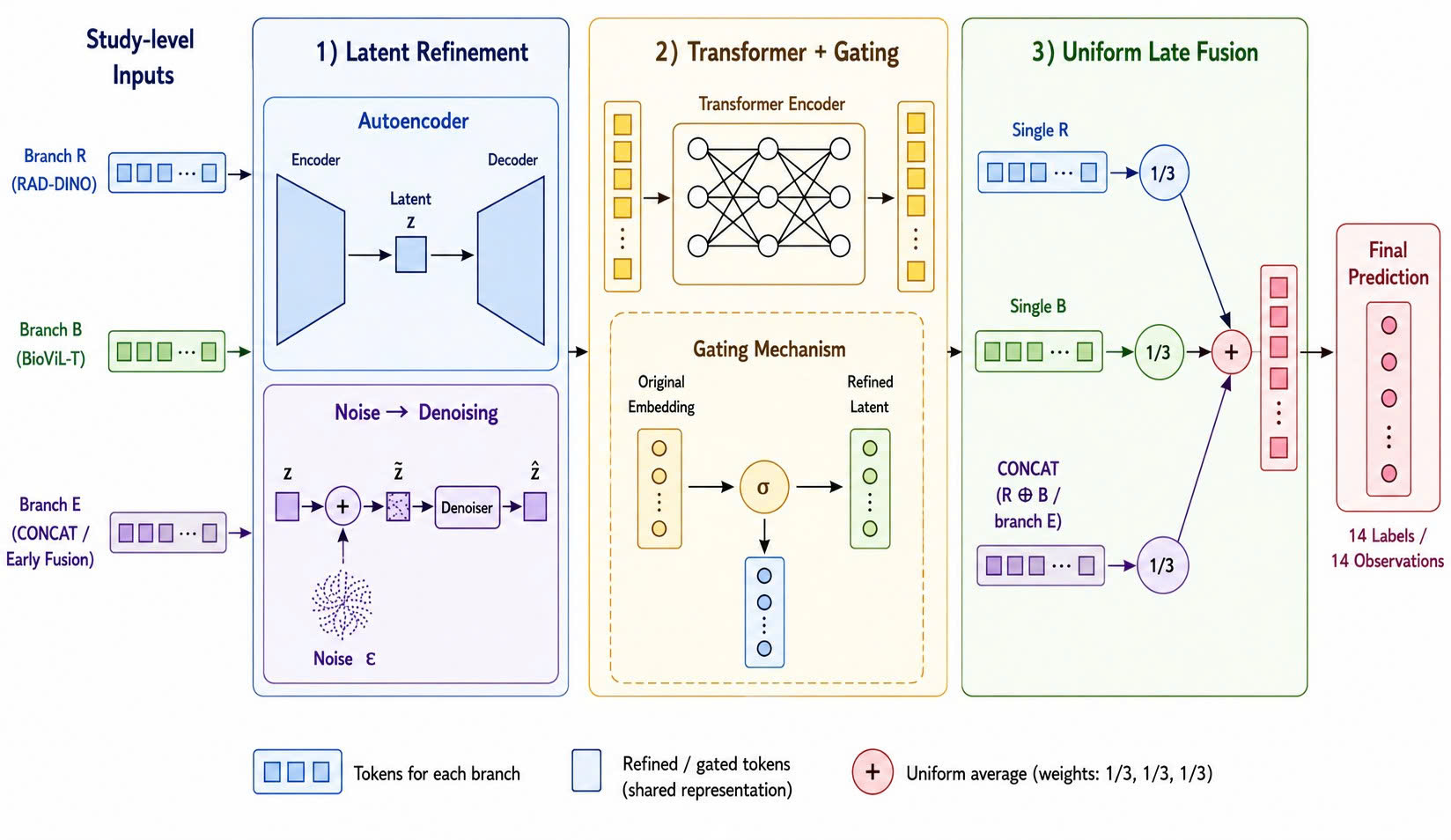}
    \caption{Overview of the latent hybrid fusion architecture.}
    \label{fig:latent_label_wise_late_fusion}
\end{figure*}

Figure~\ref{fig:latent_label_wise_late_fusion} summarizes the architecture,
which contains RAD-DINO, BioViL-T, and early-fusion branches, denoted by
\(R\), \(B\), and \(E\), respectively.

The branch inputs and latent encodings are
\begin{equation}
\begin{gathered}
    x_{R}=x_{\mathrm{RAD}},
    \qquad
    x_{B}=x_{\mathrm{Bio}},
    \qquad
    x_{E}
    =
    \left[
        x_{\mathrm{RAD}};
        x_{\mathrm{Bio}}
    \right],
    \\[3pt]
    z_{m}
    =
    f_{\mathrm{enc}}^{(m)}(x_{m}),
    \qquad
    z_{m}\in\mathbb{R}^{256},
    \qquad
    m\in\{R,B,E\}.
\end{gathered}
\label{eq:branch_latent_encoding}
\end{equation}

Each standardized embedding is mapped by a separate autoencoder to a common
256-dimensional latent space, from input dimensions 768, 128, and 896 for
RAD-DINO, BioViL-T, and early fusion, respectively. The autoencoder is trained
with Smooth L1 reconstruction loss and Gaussian input noise with standard
deviation \(0.025\). A branch-specific latent denoiser is then trained with
\(\epsilon\sim\mathcal{N}(0,I)\) and \(s\sim\mathcal{U}(0,0.65)\):
\begin{equation}
\begin{aligned}
    \widetilde{z}_{m}
    &=
    z_{m}+s\epsilon,\\
    \widehat{z}_{m}
    &=
    g_{\theta_m}
    \left(
        \widetilde{z}_{m},s
    \right),\\
    \mathcal{L}_{\mathrm{denoise}}^{(m)}
    &=
    \left\|
        \widehat{z}_{m}-z_{m}
    \right\|_{2}^{2}.
\end{aligned}
\label{eq:latent_denoising}
\end{equation}

At inference, no additional corruption is applied and the clean latent is
transformed as
\(\widehat{z}_{m}=g_{\theta_m}(z_m,0)\).
Sensitivity to the noise ceiling and the inference-time transformation is
examined in Appendices D.2 and D.3.

Each branch classifier uses both the original embedding
and the refined latent representation. They are projected to two 512-dimensional tokens, processed
jointly by a Transformer Encoder, and combined with a parallel learned gate.
The Transformer outputs are mean-pooled, while the gate learns nonnegative
weights \(\alpha_{x,m}\) and \(\alpha_{z,m}\) satisfying
\(\alpha_{x,m}+\alpha_{z,m}=1\). The resulting branch representation and
logits are
\begin{equation}
\begin{aligned}
    t_{x,m}
    &=
    W_{x,m}x_{m}, \\
    t_{z,m}
    &=
    W_{z,m}\widehat{z}_{m},\\
    h_{m}
    &=
    \frac{1}{2}h_{\mathrm{Transformer},m}
    +
    \frac{1}{2}
    \left(
        \alpha_{x,m}t_{x,m}
        +
        \alpha_{z,m}t_{z,m}
    \right),\\
    l_{m}
    &=
    \operatorname{head}_{m}(h_{m})
    \in\mathbb{R}^{14}.
\end{aligned}
\label{eq:branch_prediction}
\end{equation}

Each branch is trained with asymmetric loss~\cite{Ridnik_2021_ICCV};
training, early stopping, and checkpoint averaging are detailed in
Appendix B. After independent branch training,
logits are normalized per label using validation-set statistics and fused as
\begin{equation}
\begin{aligned}
    \overline{l}_{m,c}
    &=
    \frac{l_{m,c}-\mu_{m,c}}{\sigma_{m,c}},
    \qquad
    m\in\{R,B,E\},\\
    l_{c}^{\mathrm{late}}
    &=
    w_{R,c}\overline{l}_{R,c}
    +
    w_{B,c}\overline{l}_{B,c}
    +
    w_{E,c}\overline{l}_{E,c}.
\end{aligned}
\label{eq:label_wise_late_fusion}
\end{equation}

Fusion weights are nonnegative and sum to one. The main configuration uses
uniform weights,
\(w_{R,c}=w_{B,c}=w_{E,c}=1/3\), for every label. For analysis only,
label-specific weights are obtained by validation-set grid search with step
size \(0.1\) and reported in Appendix C.2.
Classification thresholds are selected independently per label. Component
contributions are evaluated in the ablation of
Section~\ref{subsec:ablation-study}.

\subsection{Statistical analysis}
\label{subsec:statistical_analysis}

{\setlength{\parskip}{0pt}

Confirmatory inference comprises six pre-specified paired comparisons:
\{Early $-$ BioViL-T, Early $-$ RAD-DINO, Hybrid $-$ Early\} on macro
AUROC and mAP. All tests are two-sided, with significance declared at
$p<0.001$ and simultaneous Bonferroni-level confidence intervals at
$99.17\%$. The hypothesis family, threshold, and metrics were fixed before
test-set analysis, although the study was not externally preregistered.
Macro F1, per-label results, secondary metrics, and comparisons with external
studies are treated as exploratory and carry no significance claims.\par

Equivalence is assessed separately using two one-sided tests (TOST) at
$\alpha=0.05$ with 90\% confidence intervals. The equivalence margins are $\pm0.0035$ macro AUROC and
$\pm0.0029$ mAP, corresponding to the smallest effect considered
meaningful in this study and were fixed before equivalence testing.

Optimization variability is assessed across five random seeds using paired
$t$-tests (Table~\ref{tab:paired-comparisons}). Sampling uncertainty is
quantified with a patient-level clustered bootstrap (\(B=2{,}000\)), in which
subjects are resampled with replacement and paired differences are computed
on matched resamples across configurations. Per-label AUROC differences are
additionally examined with the DeLong method as an exploratory analytic
check. Effect sizes are reported on the original metric scale, together with
relative gains and, for AUROC, reduction of the remaining ranking-error
headroom.

}

\section{Data processing}
\label{sec:data_processing}

\subsection{Study-level multi-view aggregation}
\label{subsec:multi_view_processing}

We apply the same multi-view aggregation rule to RAD-DINO and BioViL-T so
that each study is represented by a single vector. Following the MIMIC-CXR
view convention, images labeled AP or PA are treated as \textit{frontal} and
those labeled LATERAL or LL as \textit{lateral}, and the embeddings within
each view group are averaged to give \(\overline{x}_{m,F}^{(i)}\) and
\(\overline{x}_{m,L}^{(i)}\), where \(m \in \{R,B\}\) denotes the RAD-DINO
or BioViL-T encoder. When a study contains both view groups, its study-level
embedding is
\begin{equation}
    x_{m}^{(i)}
    =
    0.6 \times \overline{x}_{m,F}^{(i)}
    +
    0.4 \times \overline{x}_{m,L}^{(i)};
\end{equation}
when only one group is present, that group's embedding is used directly. The
frontal view receives the larger weight because it is the primary view for
observing the lung fields as a whole; the 0.6/0.4 weighting is selected on the
validation set from a sensitivity analysis over candidate aggregation schemes,
reported in Appendix D.1.

\subsection{Cohort construction and label processing}
\label{subsec:cohort_construction}
The input images for RAD-DINO and BioViL-T are obtained from MIMIC-CXR-JPG,
with metadata linked to embeddings from the two encoders through \texttt{subject\_id} and \texttt{study\_id}. After linkage, the final cohort contains \(211{,}130\) studies and achieves \(100\%\) coverage for both embedding sources. The train--validation--test partition follows the temporal split defined in prior work, preserving chronological order across studies and preventing future studies from being used to train models evaluated on earlier studies. This design mirrors the intended clinical use of such a model: at deployment time, ground-truth labels are available only for past studies, and the model must generalize forward in time to studies it has never observed. We note that random or ratio-based partitions of MIMIC-CXR
remain common practice, including in models reporting state-of-the-art
results~\cite{li2024mbranet}. Each study is assigned a binary label vector \(\mathbf{y}^{(i)} \in \{0,1\}^{14}\), corresponding to 14 clinical observations. \textit{Uncertain} labels are handled using the U-Ones strategy: positive and uncertain values are mapped to \(1\), whereas negative or unreported values are mapped to \(0\). The label distribution in the cohort is presented in Appendix A.

Before being passed to the encoders, each image is cropped to
\(559 \times 559\) and resized to \(518 \times 518\). RAD-DINO and BioViL-T
produce representations \(x_{R}^{(i)} \in \mathbb{R}^{768}\) and
\(x_{B}^{(i)} \in \mathbb{R}^{128}\), respectively. Both encoders
are initialized with pretrained weights and remain frozen throughout
downstream training. For BioViL-T, the study uses only the current CXR image
to extract embeddings; prior images and temporal information are not provided
to the encoder. After processing, each study is represented by a
768-dimensional RAD-DINO embedding, a 128-dimensional BioViL-T embedding, an
896-dimensional early-fusion representation, and a 14-dimensional label
vector. Details of the cohort, train--validation--test split, random seeds, and
hyperparameters are provided in Appendix B.

\section{Experimental results}
\subsection{Overall performance}
\label{subsec:overall}

Table~\ref{tab:comparison-densenet} presents study configurations (i), (ii),
(iii), and (v) across five seeds; values below are reported as the mean with
the \(95\%\) confidence interval across seeds in brackets. Of the two
configurations using a single embedding source, RAD-DINO outperforms BioViL-T
on both metrics, with mean AUROCs of \(0.827\) \([0.825, 0.828]\) and
\(0.805\) \([0.804, 0.807]\), respectively, and early fusion of the two
sources improves on either alone, indicating that they contain complementary
features. Hybrid fusion is the strongest configuration, reaching a mean AUROC
of \(0.840\) \([0.839, 0.842]\) and an mAP of \(0.467\) \([0.462, 0.471]\).
Relative to early fusion it gains \(0.0035\) AUROC and \(0.0029\) mAP, both
significant at the fixed threshold of \(p<0.001\)
(Table~\ref{tab:paired-comparisons}), so the benefit of hybrid fusion is
consistent across the experimental seeds.

The patient-level clustered bootstrap confirms these conclusions under
sampling uncertainty: for all six confirmatory comparisons, the simultaneous
\(99.17\%\) intervals of the paired differences exclude zero
(Table 7, Appendix C.1). The gain of
hybrid fusion over early fusion has a \(99.17\%\) lower bound of \(0.0030\)
and is equivalent to a \(2.12\%\) reduction of the remaining ranking-error
headroom, and the fusion-versus-single-source gains are substantially larger.
Macro F1 is treated as exploratory and moves in the same direction for all
three comparisons.

Per-label AUROC differences are additionally tested with the DeLong method on
a one-study-per-patient subset (Table 8, Appendix C.1). These rates are descriptive and carry no
confirmatory significance claims: per-label power is limited for the
hybrid-versus-early comparison, which is also the smallest of the three
effects, and confirmatory inference is therefore placed at the macro level.

\begin{table*}[t]
\centering
\caption{Comparison with the DenseNet121 results reported by
Singh~\cite{singh2024computer}.}
\label{tab:comparison-densenet}
\begin{threeparttable}
\footnotesize
\renewcommand{\arraystretch}{1.16}
\begin{tabular*}{\linewidth}{
    @{\extracolsep{\fill}}
    l
    c
    c
    c
    @{}
}
\toprule
\textbf{Configuration}
& \textbf{AUROC (95\% CI)}
& \(\boldsymbol{\Delta}\) \textbf{AUROC}
& \textbf{mAP (95\% CI)} \\
\midrule
DenseNet121~\cite{singh2024computer}
& 0.773 
& --
& -- \\
RAD-DINO only
& 0.827 [0.825, 0.828]
& 0.053
& 0.443 [0.439, 0.448] \\
BioViL-T only
& 0.805 [0.804, 0.807]
& 0.032
& 0.408 [0.404, 0.411] \\
Early fusion
& 0.837 [0.835, 0.838]
& 0.064
& 0.464 [0.459, 0.469] \\
Hybrid fusion
& \textbf{0.840} [0.839, 0.842]
& \textbf{0.067}
& \textbf{0.467} [0.462, 0.471] \\
\bottomrule
\end{tabular*}
\begin{tablenotes}[flushleft]
\scriptsize
\item Note: Results for the configurations in this study are presented as the
mean and 95\% confidence interval across five seeds. \(\Delta\) AUROC denotes
the difference from the unrounded DenseNet121 AUROC (\(0.7733\)). Because
the reference study does not report mAP, the corresponding value is denoted by
``--''. The best mean value among the study configurations is shown in bold.
\end{tablenotes}
\end{threeparttable}
\end{table*}

\begin{table*}[t]
\centering
\caption{Statistical tests across experiments.}
\label{tab:paired-comparisons}
\begin{threeparttable}
\scriptsize
\renewcommand{\arraystretch}{1.18}
\begin{tabular*}{\linewidth}{
    @{\extracolsep{\fill}}
    l
    l
    c c c
    c c c
    @{}
}
\toprule
\textbf{Model}
& \textbf{Reference model}
& \(\boldsymbol{\Delta}\) \textbf{AUROC}
& \(\boldsymbol{p}\)
& \textbf{Sig.}
& \(\boldsymbol{\Delta}\) \textbf{mAP}
& \(\boldsymbol{p}\)
& \textbf{Sig.} \\
\midrule
Early fusion
& BioViL-T only
& 0.031
& \(3.61\times10^{-9}\)
& \(\checkmark\)
& 0.056
& \(7.51\times10^{-8}\)
& \(\checkmark\) \\
Early fusion
& RAD-DINO only
& 0.010
& \(1.01\times10^{-6}\)
& \(\checkmark\)
& 0.020
& \(2.54\times10^{-6}\)
& \(\checkmark\) \\
Hybrid fusion
& Early fusion
& 0.003
& \(6.30\times10^{-5}\)
& \(\checkmark\)
& 0.003
& \(3.28\times10^{-4}\)
& \(\checkmark\) \\
\bottomrule
\end{tabular*}
\begin{tablenotes}[flushleft]
\scriptsize
\item Note:
\(\Delta\) is calculated as the result of the model in the first column minus
that of the reference model in the second column. The \(p\)-values are
computed using two-sided paired \(t\)-tests across five seeds. \textit{Sig.}
denotes a statistically significant difference at \(p<0.001\), which is
stricter than the Bonferroni-corrected level for the six pre-specified
comparisons.
\end{tablenotes}
\end{threeparttable}
\end{table*}

\subsection{Comparison with previous methods}
\label{subsec:comparison_other_models}
Table~\ref{tab:comparison_other_models} provides a reference comparison
between the proposed method and multi-label classification models evaluated
on MIMIC-CXR or MIMIC-CXR-JPG. Most of the compared methods do not release
their source code, so their results cannot be reproduced under our cohort and
evaluation protocol; except for ADNet, which we reproduced from its official
implementation, all reference values are quoted as reported in the original
papers. Because the studies do not use identical numbers of labels, cohorts,
or evaluation protocols, the reported results serve as a reference only and
may not be fully comparable, and we do not draw conclusions about the
proposed method from this table beyond noting that its performance is
competitive.

Across all 14 labels, the proposed method achieves an AUROC of \(0.840\), an
mAP of \(0.467\), and a Macro F1 of \(0.484\), the highest values among the
listed results. This finding indicates that combining RAD-DINO and BioViL-T
provides competitive performance on the 14-observation MIMIC-CXR-JPG
classification task. For a more appropriate comparison with MCX-Net, which is evaluated on 13 labels excluding \textit{No Finding}, we recalculate the metrics on the same
label set. Ours\textsuperscript{13} achieves an AUROC of \(0.837\) and a Macro
F1 of \(0.464\), exceeding MCX-Net by \(0.021\) and \(0.013\), respectively.
Notably, MCX-Net additionally uses clinical history at inference time, whereas
the proposed method uses only image representations. The result shows that the
model remains competitive in an image-only-at-inference setting.

\begin{table*}[t]
\centering
\caption{Comparison of the proposed method with other models on MIMIC-CXR.}
\label{tab:comparison_other_models}
\small
\renewcommand{\arraystretch}{1.10}
\setlength{\tabcolsep}{8pt}
\begin{tabularx}{0.88\linewidth}{
    @{}
    >{\raggedright\arraybackslash}X
    S[table-format=1.3]
    S[table-format=1.3]
    S[table-format=1.3]
    @{}
}
\toprule
{\textbf{Model}}
& {\textbf{AUROC} $\uparrow$}
& {\textbf{mAP} $\uparrow$}
& {\textbf{Macro F1} $\uparrow$}
\\
\midrule
\multicolumn{4}{@{}l}{\textit{Comparison on 13 labels (excluding No Finding)}} \\[1pt]
MCX-Net\textsuperscript{13}~\cite{yang2025enhancing}
& 0.816
& \multicolumn{1}{c}{--}
& 0.451
\\
Ours\textsuperscript{13}
& {\bfseries 0.837}
& \multicolumn{1}{c}{--}
& {\bfseries 0.464}
\\
\addlinespace[4pt]
\cmidrule{1-4}
\multicolumn{4}{@{}l}{\textit{Comparison on 14 labels}} \\[1pt]
DenseNet121~\cite{singh2024computer}
& 0.773
& \multicolumn{1}{c}{--}
& \multicolumn{1}{c}{--}
\\
MBRANet~\cite{li2024mbranet}
& 0.805
& \multicolumn{1}{c}{--}
& \multicolumn{1}{c}{--}
\\
Jackson et al.~\cite{jackson2024fairness}
& 0.819
& \multicolumn{1}{c}{--}
& \multicolumn{1}{c}{--}
\\
ADNet~\cite{KANG2024108198}\textsuperscript{\dag}
& 0.829
& 0.422
& 0.433
\\
Ours\textsuperscript{14}
& {\bfseries 0.840}
& {\bfseries 0.467}
& {\bfseries 0.484}
\\
\bottomrule
\end{tabularx}
\vspace{3pt}
\begin{minipage}{0.88\linewidth}
\footnotesize
\textit{Note:}
MCX-Net\textsuperscript{13} uses both CXR images and clinical history at
inference time and is evaluated on 13 labels, excluding \textit{No Finding}.
For a direct comparison, Ours\textsuperscript{13} is likewise calculated on
the same 13 labels after excluding \textit{No Finding}, whereas
Ours\textsuperscript{14} denotes the results on all 14 labels. Results marked
with \textsuperscript{\dag} were reproduced using the official source code
released by the authors.
\end{minipage}
\end{table*}

\subsection{Per-label results}
\label{subsec:label_results}
\begin{figure}[t]
    \centering
    \includegraphics[width=\columnwidth,keepaspectratio]{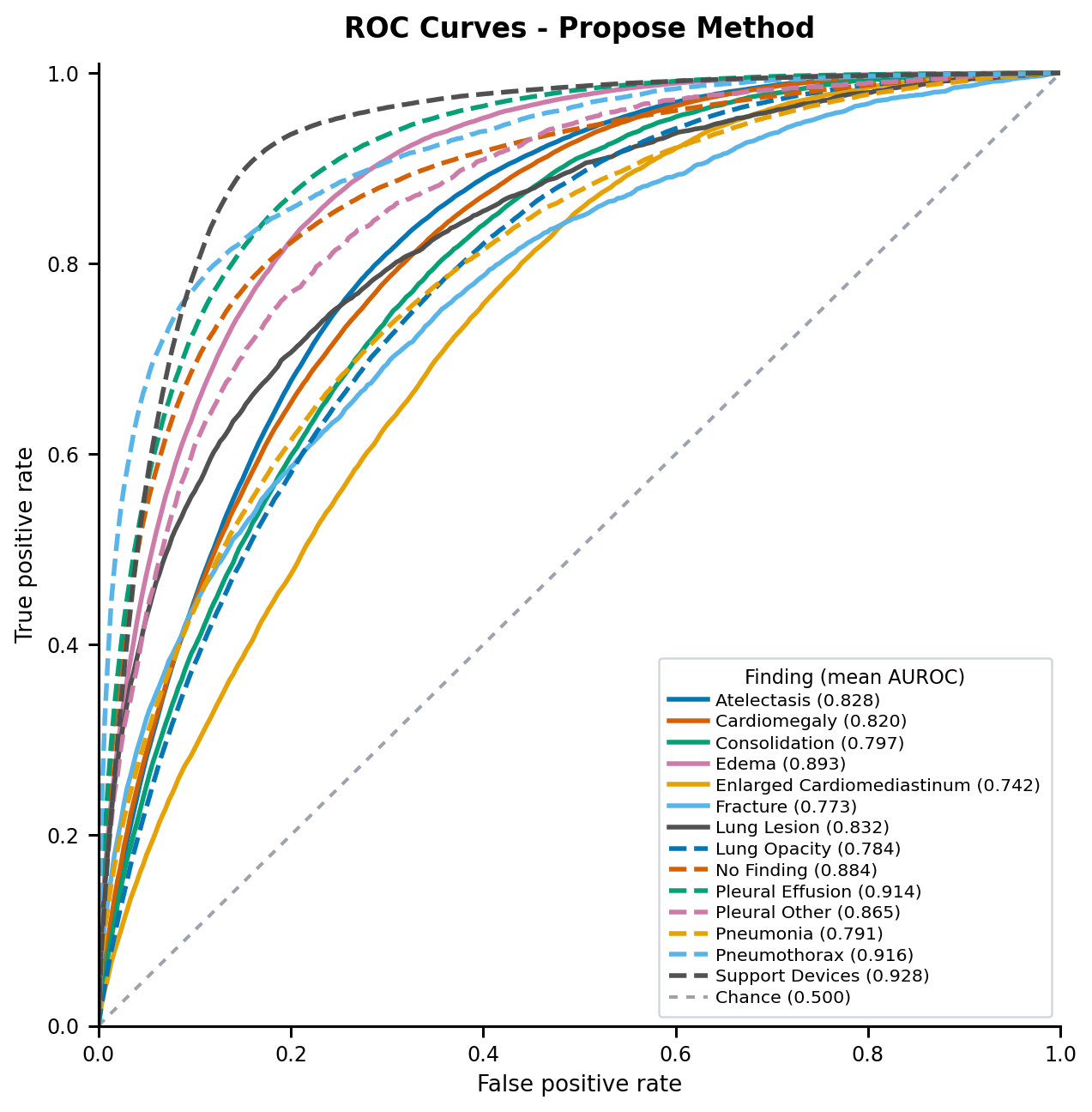}
    \caption{ROC curves of latent hybrid fusion for the 14 labels on the
    test set. Mean AUROC ranges from $0.742$ for \textit{Enlarged
    Cardiomediastinum} to $0.928$ for \textit{Support Devices}.}
    \label{fig:roc-curves-labels}
\end{figure}

Figure~\ref{fig:roc-curves-labels} presents the ROC curves of our method for
the 14 labels; complete per-label metrics and the detailed comparison with
MBRANet are provided in Figure 1 and Table 6, respectively, in
Appendix C.1. \textit{Support Devices},
\textit{Pneumothorax}, and \textit{Pleural Effusion} obtain the highest
AUROCs, and \textit{Support Devices} and \textit{Pleural Effusion} also rank
highest on AP. In contrast, \textit{Pleural Other}, \textit{Fracture}, and
\textit{Enlarged Cardiomediastinum} have substantially lower AP values.
\textit{Pleural Other} illustrates the gap most clearly, with an AUROC of
\(0.865\) but an AP of only \(0.104\), showing that a relatively high AUROC
does not necessarily reflect strong positive-class performance when the label
distribution is imbalanced.

In the per-label comparison with MBRANet, the proposed method has the higher
macro AUROC and performs better on 10 of the 14 labels, with the largest gain
on \textit{Pneumothorax} (\(+0.147\)). However, it still underperforms MBRANet
on several labels, most notably \textit{Enlarged Cardiomediastinum}
(\(-0.081\)). The improvement is therefore broad but not uniform across
labels.

\subsection{Ablation Study}
\label{subsec:ablation-study}

To localize the source of the observed improvements, we evaluate eleven configurations (A1--A11; Table~\ref{tab:ablation-study}) covering latent refinement, prediction head, and fusion under the same experimental protocol and five seeds as the main experiments. Pairwise comparisons were specified before inspection of the ablation results (Section~\ref{subsec:statistical_analysis}), with significance declared at \(p<0.001\).\par
\noindent\textit{Latent refinement (A1--A3).} The autoencoder alone does not significantly improve over A1, and adding denoising on top of the autoencoder yields no significant incremental gain; only the complete latent-refinement block improves over A1. We therefore interpret the autoencoder and noisy-latent training jointly rather than as independently beneficial components. Sensitivity analyses further show that removing the denoising objective leaves performance unchanged, whereas bypassing the learned inference-time transformation removes the block-level gain (Appendices D.2 and D.3). These results support a block-level interpretation in which the learned nonlinear transformation, rather than the denoising objective itself, is the operative component.\par
\noindent\textit{Prediction head (A4--A7).} With latent refinement fixed, the MLP, Transformer-only, gating-only, and Transformer-plus-gating heads show no significant pairwise differences. We therefore find no evidence that prediction-head choice explains the observed gain. This does not establish equivalence, which the five-seed design was not powered to test.\par
\noindent\textit{Fusion stage (A8--A11).} Uniform fusion of the refined branches (A8) performs best, whereas learned label-wise fusion without latent refinement (A9) and the capacity-matched CONCAT--MLP control (A10) perform significantly worse. Because A9 changes both representation quality and fusion rule, the factorial analysis in Appendix E.1 separates these effects and shows that refined representations are the dominant factor. Learned weights help at the raw level but provide no significant additional benefit once the branches are refined.\par
\noindent A deep ensemble (A11) does not significantly differ from A8. TOST against the pre-specified equivalence margins further places the two configurations within the smallest effect considered meaningful in this study (Appendix E.2), supporting an ensemble-effect interpretation of the fusion stage.\par
\noindent Overall, the ablation attributes the gain to latent refinement and the combination of complementary embedding sources, with an additional ensemble effect at fusion, rather than to prediction-head choice, added capacity, or a more complex fusion rule. Calibration of these configurations after per-label Platt scaling is reported in Appendix C.4; all of them are well calibrated, and such analysis is exploratory.

\begin{table*}[t]
\centering
\caption{Ablation over the latent-refinement block, the prediction head, and
the fusion stage. Each value is the mean macro AUROC and 95\% confidence
interval across five seeds; $p$ is the two-sided paired $t$-test p-value
across seeds.}
\label{tab:ablation-study}
\begin{threeparttable}
\scriptsize
\renewcommand{\arraystretch}{1.18}
\begin{tabular*}{\linewidth}{
    @{\extracolsep{\fill}}
    l
    l
    c
    c
    c
    c
    c
    @{}
}
\toprule
\textbf{ID} & \textbf{Configuration} & \textbf{AUROC (95\% CI)}
& \textbf{Comparison} & \(\boldsymbol{\Delta}\) \textbf{AUROC}
& \(\boldsymbol{p}\) & \textbf{Sig.} \\
\midrule
\multicolumn{7}{@{}l}{\textit{Latent-refinement block}} \\
A1  & Embedding only
    & 0.8356 [0.8340, 0.8372] & & & & \\
A2  & A1 + Autoencoder
    & 0.8366 [0.8346, 0.8386] & A2 $-$ A1  & $+0.0010$
    & \(2.74\times10^{-2}\) & $\times$ \\
A3  & A2 + noise
    & 0.8369 [0.8354, 0.8385] & A3 $-$ A1  & $+0.0013$
    & \(1.64\times10^{-4}\) & \checkmark \\
\midrule
\multicolumn{7}{@{}l}{\textit{Prediction head (Latent-refinement gain under different heads)}} \\
A4  & MLP head
    & 0.8368 [0.8352, 0.8384] & A4 $-$ A1  & $+0.0012$
    & \(1.68\times10^{-2}\) & $\times$ \\
A5  & Transformer only
    & 0.8369 [0.8354, 0.8385] & A5 $-$ A1  & $+0.0013$
    & \(4.84\times10^{-3}\) & $\times$ \\
A6  & Gating only
    & 0.8364 [0.8346, 0.8383] & A6 $-$ A1  & $+0.0008$
    & \(4.98\times10^{-2}\) & $\times$ \\
A7  & Transformer + gating
    & 0.8369 [0.8354, 0.8385] & A7 $-$ A1  & $+0.0013$
    & \(1.64\times10^{-4}\) & \checkmark \\
\midrule
\multicolumn{7}{@{}l}{\textit{Fusion stage}} \\
A8  & Uniform fusion
    & 0.8404 [0.8388, 0.8419] & & & & \\
A9  & Learned fusion, no latent
    & 0.8364 [0.8352, 0.8376] & A9 $-$ A8  & $-0.0040$
    & \(4.20\times10^{-5}\) & \checkmark \\
A10 & CONCAT--MLP
    & 0.8364 [0.8343, 0.8385] & A10 $-$ A8 & $-0.0040$
    & \(9.62\times10^{-4}\) & \checkmark \\
A11 & Deep ensemble
    & 0.8399 [0.8382, 0.8416] & A11 $-$ A8 & $-0.0005$
    & \(1.55\times10^{-1}\) & $\times$ \\
\bottomrule
\end{tabular*}
\begin{tablenotes}[flushleft]
\scriptsize
\item Note: A \(\checkmark\) denotes a statistically significant difference,
whereas \(\times\) denotes a nonsignificant difference, at the threshold
\(p<0.001\). The sign of \(\Delta\) indicates the direction of the
difference. A7 is identical to A3, so A7 $-$ A1 coincides with A3 $-$ A1.
A8 is the main configuration, whose overall results are reported in
Section~\ref{subsec:overall}. Configuration details are described in the text, and configuration A9
differs from A8 in two respects simultaneously: the branch representations
and the fusion rule. Appendix E.1 decomposes this
contrast.
\end{tablenotes}
\end{threeparttable}
\end{table*}

\subsection{Performance by Label Prevalence}
\label{subsec:prevalence}
To examine whether the fusion benefit depends on label frequency, we stratify the 14 labels into common ($>15\%$, 8 labels), mid-prevalence ($5$--$15\%$, 3 labels), and rare ($<5\%$, 3 labels) tiers according to the prevalence in Appendix A, and compare hybrid fusion against early fusion within each tier using per-label bootstrap confidence intervals of the improvement. Table~\ref{tab:prevalence_tiers} reports the mean improvement and the
number of labels whose interval excludes zero in each tier: the improvement
is well supported on common labels for both metrics, and mid-prevalence labels show a similar but less uniform pattern. On the three rare labels, the evidence is asymmetric across metrics: the AUROC improvements are supported for all three labels, whereas none of the AP intervals excludes zero, reflecting the small number of positive studies available for these labels. The supported conclusion is therefore ranking-level: fusion improves the ordering of studies on the rare labels, and the effect on average
precision remains undetermined. This ranking-level pattern is consistent
with the branch-level error structure reported in Appendix C.3. The labels with the lowest between-branch error correlation are \textit{Fracture} (0.817), \textit{Pleural Other} (0.796), \textit{Pneumothorax} (0.816), and \textit{Lung Lesion} (0.855), against a macro average of 0.918. Three of the four are rare labels, and the fourth, Pneumothorax, is the least prevalent mid-tier label; these labels retain the most residual disagreement for fusion to exploit.

\begin{table}[!t]
\centering
\caption{Improvement of hybrid fusion over early fusion by label-prevalence
tier (exploratory).}
\label{tab:prevalence_tiers}
\footnotesize
\renewcommand{\arraystretch}{1.16}
\setlength{\tabcolsep}{8pt}
\resizebox{\columnwidth}{!}{%
\begin{tabular}{lccccc}
\toprule
\textbf{Tier} & \textbf{Labels}
& \(\boldsymbol{\Delta}\)\textbf{AUROC} & \textbf{Excl.\ 0}
& \(\boldsymbol{\Delta}\)\textbf{AP} & \textbf{Excl.\ 0} \\
\midrule
Common ($>15\%$)  & 8 & \(+0.0022\) & 8/8 & \(+0.0038\) & 8/8 \\
Mid ($5$--$15\%$) & 3 & \(+0.0036\) & 3/3 & \(+0.0024\) & 2/3 \\
Rare ($<5\%$)     & 3 & \(+0.0067\) & 3/3 & \(+0.0012\) & 0/3 \\
\bottomrule
\end{tabular}
}
\end{table}

\section{Discussion}
\label{sec:discussion}
\subsection{Effectiveness of RAD-DINO representations on MIMIC-CXR}

RAD-DINO provides a strong frozen visual representation for multi-label CXR
classification in our experimental setting. Despite using only a single global
embedding as input to the downstream classifier, the RAD-DINO-only branch
achieves competitive performance without fine-tuning the pretrained encoder.
This suggests that image-only biomedical pretraining provides a useful
representation for the downstream task. Cross-study comparisons with prior
methods are treated only as contextual references because their cohorts and
evaluation protocols differ from ours.

\subsection{Complementarity between RAD-DINO and BioViL-T}

{\setlength{\parskip}{0pt}

The results show that BioViL-T underperforms RAD-DINO when used independently,
with differences of \(0.021\) AUROC and \(0.036\) mAP. However, lower
standalone performance does not imply that BioViL-T representations contain no
complementary information. Directly concatenating the two embeddings through
early fusion raises AUROC from \(0.827\) to \(0.837\) and mAP from
\(0.443\) to \(0.464\) relative to RAD-DINO, and these improvements are
statistically significant.\par

This result shows that the two representation sources are not entirely
redundant. RAD-DINO is learned from unimodal visual signals, whereas BioViL-T's
image representation is formed under vision--language supervision. A
representation-similarity analysis (Appendix C.3)
characterizes this relationship: linear CKA between the two embeddings is
\(0.419\), far above a permutation baseline of \(\approx 0.002\) and far
below the value of 1 attained by linearly equivalent representations, yet
CKA rises to \(0.834\) between the refined latents and the error
correlation of the two branch predictions reaches a macro Pearson of
\(0.918\). Training on the same labels thus drives the two branches toward
similar prediction functions, which explains why the additional gain from
fusing them is small even though the underlying representations differ.
Although BioViL-T is not as strong as RAD-DINO on its own, these
complementary signals can still improve the joint representation when the
embeddings are combined. This supports the hypothesis that purely visual
and vision--language representations are complementary for multi-label
chest X-ray classification.

}

\subsection{Role of latent space}
RAD-DINO, BioViL-T, and CONCAT differ in dimensionality, pretraining objective, and feature distribution; predictors built from the three embedding sources may therefore provide non-equivalent signals. The ablation shows that the autoencoder applied alone yields no significant improvement, and that late fusion of unrefined branches falls significantly below the fusion of refined branches.

In contrast, the best performance occurs when each embedding source is transformed in a separate latent branch before late fusion. Appendix C.3 shows that each of the three branches contributes a unique predictive signal within the ensemble; because all branches derive from the same two embedding sources, this reflects functional diversity among the predictors. This finding shows that the method's benefit arises from maintaining separate representation sources, refining them before prediction, and combining their predictions. The eleven-configuration ablation in Section~\ref{subsec:ablation-study} further localizes this benefit to the latent-refinement block, together with an ensemble effect at the fusion stage.

\subsection{Practical implications}
One practical advantage of this setup is that both foundation encoders remain
frozen, so downstream optimization focuses only on the embedding-processing
modules and classifiers. This reduces training cost relative to fine-tuning
the entire backbone and allows pretrained representations to be reused across
different downstream configurations.

Furthermore, the system does not require radiology reports or clinical
variables at inference time. Although BioViL-T receives vision--language
supervision during pretraining, the final model still operates in an
image-only-at-inference setting.

\subsection{Limitations}
Although the proposed hybrid fusion improves on both single-source branches, on early fusion, and on prior work, this study has several limitations. First, the MIMIC-CXR-JPG labels are
extracted automatically from radiology reports and are strongly imbalanced, so
the macro metrics reported here are more favorable than performance on the
rarest labels alone; reducing each study to a single global vector may likewise
discard small or localized findings. Second, the across-seed paired $t$-tests
rest on five seeds, so their power is limited, while the patient-level
clustered bootstrap resamples subjects with the seeds held fixed and therefore
does not propagate optimization variability. Third, the main comparisons, the
ablation, and the auxiliary analyses share a single held-out test set; the
confirmatory family is restricted and the remainder labeled exploratory, yet
repeated use of one test set can still inflate the apparent consistency of the
exploratory findings, and the analysis plan was fixed internally rather than
preregistered. Finally, all experiments are conducted on MIMIC-CXR-JPG, so
generalizability under domain shift remains to be confirmed.


\section{Conclusion}

This study proposes a multi-label chest X-ray classification framework that
combines unimodal representations from RAD-DINO with vision--language
representations from BioViL-T. The two encoders remain frozen, while their
embeddings are refined in separate latent branches and fused across the three
branches. This design aims to exploit differences in the models' pretraining
information sources instead of relying on a single representation.

Results on the 14 MIMIC-CXR-JPG labels show that RAD-DINO outperforms
BioViL-T when used independently, although combining the two embeddings still
produces a clear improvement over either individual branch. Latent hybrid
fusion achieves the best performance, with a mean AUROC of \(0.840\) and an
mAP of \(0.467\). Notably, applying the same late fusion to unrefined branches performs significantly worse than the main configuration. This
finding shows that the method's effectiveness arises from the quality of the
representation supplied by each branch, while the fusion stage itself adds an
ensemble effect. However, the current results are based solely on internal
evaluation with MIMIC-CXR-JPG, and validation using data from other
healthcare institutions is needed to confirm the method's generalizability.
\subsection*{Acknowledgement}
We acknowledge Ho Chi Minh City University of Technology (HCMUT), VNU - HCM for supporting this study.

{
    \small
    \bibliographystyle{ieeenat_fullname}
    \bibliography{references}
}

\clearpage

\end{document}


\maketitle
\appendix

\renewcommand{\topfraction}{0.92}
\renewcommand{\bottomfraction}{0.85}
\renewcommand{\textfraction}{0.06}
\renewcommand{\floatpagefraction}{0.85}
\renewcommand{\dbltopfraction}{0.92}
\renewcommand{\dblfloatpagefraction}{0.85}
\setcounter{topnumber}{4}
\setcounter{bottomnumber}{3}
\setcounter{totalnumber}{6}
\setcounter{dbltopnumber}{4}


\section{Label Distribution}
\label{app:label_distribution}
\begin{table}[H]
    \centering
    \caption{Label distribution in the aligned cohort.}

    \small
    \renewcommand{\arraystretch}{1.12}
    \setlength{\tabcolsep}{8pt}

    \begin{tabularx}{0.82\linewidth}{
        @{}
        >{\raggedright\arraybackslash}X
        >{\raggedleft\arraybackslash}p{0.22\linewidth}
        >{\raggedleft\arraybackslash}p{0.20\linewidth}
        @{}
    }
        \toprule
        \textbf{Label}
        & \textbf{Positive studies}
        & \textbf{Prevalence} \\
        \midrule
        Atelectasis                    & \(52{,}818\) & \(25.017\%\) \\
        Cardiomegaly                   & \(47{,}579\) & \(22.535\%\) \\
        Consolidation                  & \(14{,}190\) & \(6.721\%\)  \\
        Edema                          & \(37{,}934\) & \(17.967\%\) \\
        Enlarged Cardiomediastinum     & \(15{,}497\) & \(7.340\%\)  \\
        Fracture                       & \(4{,}590\)  & \(2.174\%\)  \\
        Lung Lesion                    & \(6{,}801\)  & \(3.221\%\)  \\
        Lung Opacity                   & \(51{,}684\) & \(24.480\%\) \\
        No Finding                     & \(69{,}204\) & \(32.778\%\) \\
        Pleural Effusion               & \(56{,}679\) & \(26.846\%\) \\
        Pleural Other                  & \(2{,}519\)  & \(1.193\%\)  \\
        Pneumonia                      & \(32{,}292\) & \(15.295\%\) \\
        Pneumothorax                   & \(10{,}791\) & \(5.111\%\)  \\
        Support Devices                & \(63{,}135\) & \(29.903\%\) \\
        \bottomrule
    \end{tabularx}

    \vspace{3pt}

    \begin{minipage}{0.78\linewidth}
        \footnotesize
        \textit{Note:} The aligned cohort has substantial label imbalance,
        with prevalence ranging from \(1.193\%\) for \textit{Pleural Other}
        to \(32.778\%\) for \textit{No Finding}.
    \end{minipage}
\end{table}

\section{Experimental Configuration}
\label{app:experimental_setup}
\begin{table}[H]
    \centering
    \caption{Cohort and reproducibility settings.}
    \label{tab:cohort_reproducibility}

    \footnotesize
    \renewcommand{\arraystretch}{1.05}
    \setlength{\tabcolsep}{10pt}

    \begin{tabular}{
        @{}
        l
        r
        @{}
    }
        \toprule
        \textbf{Item} & \textbf{Value} \\
        \midrule
        Aligned studies
            & \(211{,}130\) \\
        Disease labels
            & \(14\) \\
        RAD-DINO embedding dimension
            & \(768\) \\
        BioViL-T embedding dimension
            & \(128\) \\
        Random seeds
            & \(42, 43, 44, 45, 46\) \\
        Number of independent runs
            & \(5\) \\
        Optimizer
            & AdamW \\
        Learning-rate scheduler
            & ReduceLROnPlateau \\
        Transformer encoder layers
            & 2 \\
        Activation Function 
            & GELU \\
        \bottomrule
    \end{tabular}
\end{table}

\vspace{30pt}

\begin{table}[H]
    \centering
    \caption{Train, validation, and test set sizes shared across all
    five random seeds.}
    \label{tab:dataset_partition}

    \footnotesize
    \renewcommand{\arraystretch}{1.05}
    \setlength{\tabcolsep}{10pt}

    \begin{tabular}{
        @{}
        l
        c
        r
        @{}
    }
        \toprule
        \textbf{Split}
            & \textbf{Ratio}
            & \textbf{Studies} \\
        \midrule
        Train
            & \(70\%\)
            & \(147{,}791\) \\
        Validation
            & \(10\%\)
            & \(21{,}113\) \\
        Test
            & \(20\%\)
            & \(42{,}226\) \\
        \midrule
        Total
            & \(100\%\)
            & \(211{,}130\) \\
        \bottomrule
    \end{tabular}
\end{table}

\vspace{30pt}

\begin{table*}[!t]
\centering
\caption{Training and model-selection hyperparameters.}
\label{tab:experimental_hyperparameters}
\footnotesize
\renewcommand{\arraystretch}{1.08}
\setlength{\tabcolsep}{6pt}
\begin{minipage}[t]{0.48\textwidth}
\begin{tabularx}{\linewidth}{
    @{}
    >{\raggedright\arraybackslash}X
    >{\raggedleft\arraybackslash}p{0.34\linewidth}
    @{}
}
    \toprule
    \textbf{Parameter} & \textbf{Setting} \\
    \midrule

    \multicolumn{2}{@{}l}{\textit{Model dimensions}} \\
    \addlinespace[2pt]

    Predictor hidden dimension
        & \(512\) \\

    Latent dimension
        & \(256\) \\
    
    Feed-forward dimension
        & \(1024\) \\ 

    \midrule
    \multicolumn{2}{@{}l}{\textit{Training schedule}} \\
    \addlinespace[2pt]

    Batch size
        & \(1024\) \\

    Autoencoder epochs
        & \(12\) \\

    Latent denoiser epochs
        & \(10\) \\

    Maximum predictor epochs
        & \(60\) \\

    Early-stopping patience
        & \(8\) \\

    Top checkpoints averaged
        & \(3\) \\

    \midrule
    \multicolumn{2}{@{}l}{\textit{Optimization}} \\
    \addlinespace[2pt]

    Autoencoder and denoiser learning rate
        & \(2.0 \times 10^{-4}\) \\

    Predictor learning rate
        & \(1.5 \times 10^{-4}\) \\

    Weight decay
        & \(3.0 \times 10^{-4}\) \\

    \midrule
    \multicolumn{2}{@{}l}{\textit{Regularization}} \\
    \addlinespace[2pt]

    Gaussian noise standard deviation
        & \(0.025\) \\

    Mixup coefficient
        & \(0.2\) \\

    Label-smoothing coefficient
        & \(0.02\) \\

    \bottomrule
\end{tabularx}
\end{minipage}
\hfill
\begin{minipage}[t]{0.48\textwidth}
\begin{tabularx}{\linewidth}{
    @{}
    >{\raggedright\arraybackslash}X
    >{\raggedleft\arraybackslash}p{0.34\linewidth}
    @{}
}
    \toprule
    \textbf{Parameter} & \textbf{Setting} \\
    \midrule
    \multicolumn{2}{@{}l}{\textit{Encoders and pooling}} \\
    \addlinespace[2pt]

    RAD-DINO checkpoint
        & \texttt{microsoft/rad-dino} \\

    BioViL-T checkpoint
        & \texttt{BiomedVLP-BioViL-T} \\

    RAD-DINO feature
        & CLS token, \(768\)-d \\

    BioViL-T feature
        & projected global embedding, \(128\)-d \\

    Within-view pooling
        & mean \\

    Frontal/lateral weighting
        & \(0.6/0.4\) \\

    \midrule
    \multicolumn{2}{@{}l}{\textit{Architecture details}} \\
    \addlinespace[2pt]

    Autoencoder dropout (enc./dec.)
        & \(0.20,\ 0.05\ /\ 0.15\) \\

    Denoiser dropout
        & \(0.20,\ 0.15\) \\

    Denoiser loss
        & MSE, \(s \sim \mathcal{U}(0, 0.65)\) \\

    Transformer attention heads
        & \(8\) \\

    Transformer dropout
        & \(0.30\) \\

    Projection / head dropout
        & \(0.10\ /\ 0.35\) \\

    Head combination
        & \(0.5\times\)pool \(+\ 0.5\times\)gated \\

    \midrule
    \multicolumn{2}{@{}l}{
        \textit{Fusion and threshold selection}
    } \\
    \addlinespace[2pt]

    Fusion weights (main configuration)
        &  \(1/3\)  \\

    Fusion grid step (for analysis)
        & \(0.1\) \\

    Shrinkage toward early-only prior
        & \(0.70\) \\

    Weight acceptance gate (val.\ \(\Delta\)AP)
        & \(\geq 0.002\) \\

    Threshold selection
        & max F1 on val.\ PR curve \\

    Threshold recall constraint
        & recall \(\geq 0.30\) \\
    \bottomrule
\end{tabularx}
\end{minipage}
\end{table*}

\begin{table}[H]
    \centering
    \caption{Inference and training cost of the fusion stage, measured
    on a single NVIDIA A100 GPU.}
    \label{tab:cost}

    \small
    \renewcommand{\arraystretch}{1.10}
    \setlength{\tabcolsep}{7pt}

    \begin{tabularx}{0.88\linewidth}{
        @{}
        >{\raggedright\arraybackslash}X
        S[table-format=2.1]
        S[table-format=2.1]
        S[table-format=1.3]
        @{}
    }
        \toprule
        {\textbf{Configuration}}
            & {\textbf{Params}}
            & {\textbf{FLOPs}}
            & {\textbf{Latency}} \\
        {}
            & {\textbf{(M)}}
            & {\textbf{(M/study)}}
            & {\textbf{(ms/study)}} \\
        \midrule
        RAD-DINO only   & 7.1  & 13.1 & 0.057 \\
        BioViL-T only   & 6.1  & 11.8 & 0.057 \\
        Early fusion    & 7.3  & 13.4 & 0.057 \\
        Hybrid fusion   & 20.5 & 38.3 & 0.170 \\
        \bottomrule
    \end{tabularx}

    \vspace{3pt}

    \begin{minipage}{0.88\linewidth}
        \footnotesize
        \textit{Note:}
        FLOPs count linear layers only and exclude the frozen encoders.
        Latency is the median per study at batch size 32. Peak VRAM is
        \(99\)\,MB at inference and \(463\)\,MB during training for all
        configurations. Hybrid fusion trains all required components in
        approximately \(4.6\) minutes per seed (about \(0.4\) GPU-hours
        for five seeds), an upper bound based on nominal epoch counts
        without early stopping. The three single-branch configurations
        have comparable capacity (\(11.8\)--\(13.4\)M FLOPs), so the
        early-fusion gain over single sources reflects added information
        rather than added capacity, whereas hybrid fusion trades
        \(2.9\times\) the compute of early fusion for the improvement
        reported in Section 5.1 of the main paper.
    \end{minipage}
\end{table}

\FloatBarrier
\section{Results}
\label{app:per_label_analysis}

\subsection{Label Results}
\label{app:label_results}
\begin{figure}[!htbp]
    \centering
    \includegraphics[
        width=\linewidth,
        keepaspectratio
    ]{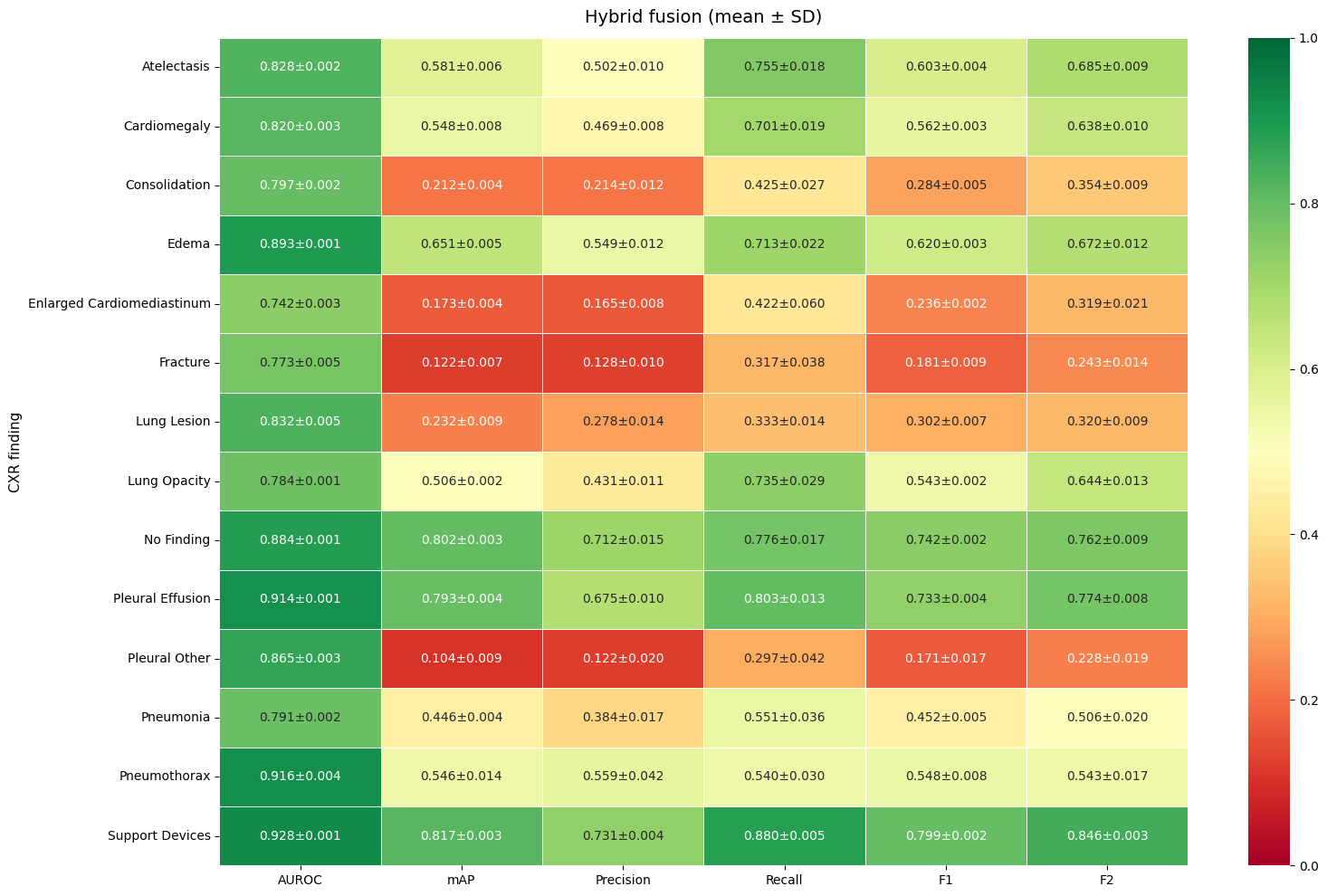}
    \captionsetup{
        justification=centering,
        singlelinecheck=false
    }
    \caption{Per-label performance of latent hybrid fusion. Each
    cell shows the mean \(\pm\) standard deviation across five seeds.}
    \label{fig:label-metrics-heatmap}
\end{figure}
\begin{table*}[t]
    \centering
    \caption{Per-label AUROC comparison between MBRANet and the proposed
    method on MIMIC-CXR.}
    \label{tab:mbranet_label_comparison}
    \small
    \renewcommand{\arraystretch}{1.10}
    \setlength{\tabcolsep}{7pt}
    \begin{tabularx}{0.82\textwidth}{
        @{}
        >{\raggedright\arraybackslash}X
        >{\centering\arraybackslash}p{0.18\textwidth}
        >{\centering\arraybackslash}p{0.18\textwidth}
        >{\centering\arraybackslash}p{0.16\textwidth}
        @{}
    }
        \toprule
        \textbf{Label}
        & \textbf{MBRANet~\cite{li2024mbranet}}
        & \textbf{Ours}
        & \(\boldsymbol{\Delta}\) \\
        \midrule
        Atelectasis
        & \(0.779\)
        & \textbf{0.828}
        & \(+0.049\) \\
        Cardiomegaly
        & \textbf{0.825}
        & \(0.820\)
        & \(-0.005\) \\
        Consolidation
        & \(0.788\)
        & \textbf{0.797}
        & \(+0.009\) \\
        Edema
        & \(0.833\)
        & \textbf{0.893}
        & \(+0.060\) \\
        Cardio.
        & \textbf{0.823}
        & \(0.742\)
        & \(-0.081\) \\
        Fracture
        & \(0.703\)
        & \textbf{0.773}
        & \(+0.070\) \\
        Lung Lesion
        & \(0.735\)
        & \textbf{0.832}
        & \(+0.097\) \\
        Lung Opacity
        & \textbf{0.788}
        & \(0.784\)
        & \(-0.004\) \\
        No Finding
        & \(0.835\)
        & \textbf{0.884}
        & \(+0.049\) \\
        Pleural Effusion
        & \(0.863\)
        & \textbf{0.914}
        & \(+0.051\) \\
        Pleural Other
        & \(0.821\)
        & \textbf{0.865}
        & \(+0.044\) \\
        Pneumonia
        & \textbf{0.823}
        & \(0.791\)
        & \(-0.032\) \\
        Pneumothorax
        & \(0.769\)
        & \textbf{0.916}
        & \(+0.147\) \\
        Support Devices
        & \(0.879\)
        & \textbf{0.928}
        & \(+0.049\) \\
        \midrule
        \textbf{Macro average}
        & \(0.805\)
        & \textbf{0.840}
        & \(+0.035\) \\
        \bottomrule
    \end{tabularx}
    \vspace{3pt}
    \begin{minipage}{0.82\textwidth}
        \footnotesize
        \textit{Note:} MBRANet results are quoted as reported in the original paper. Bold values indicate the higher AUROC for each label.
        \(\Delta\) is calculated as the AUROC of the proposed method minus
        that of MBRANet.
    \end{minipage}
\end{table*}


\begin{table*}[!t]
\centering
\caption{Patient-level clustered bootstrap of paired metric differences ($B = 2{,}000$ subject-level resamples; within each replicate the paired difference is computed per seed on an independently drawn subject resample and averaged across the five seeds). Confirmatory comparisons report simultaneous $99.17\%$ confidence intervals; Macro F1 is exploratory and reports $95\%$ intervals.}
\label{tab:bootstrap_ci}
\begin{threeparttable}
\scriptsize
\renewcommand{\arraystretch}{1.18}
\begin{tabular*}{\textwidth}{
    @{\extracolsep{\fill}}
    l l c c c c
    @{}
}
\toprule
\textbf{Comparison} & \textbf{Metric} & \(\boldsymbol{\Delta}\)
& \textbf{CI} & \textbf{Rel.\ gain (\%)} & \textbf{Headroom (\%)} \\
\midrule
Early $-$ BioViL-T & AUROC & 0.0314 & [0.0301, 0.0327] & 3.90 & 16.15 \\
                   & mAP   & 0.0560 & [0.0544, 0.0579] & 13.75 & -- \\
                   & Macro F1\textsuperscript{e} & 0.0451 & [0.0437, 0.0464] & 10.33 & -- \\
\addlinespace[2pt]
Early $-$ RAD-DINO & AUROC & 0.0104 & [0.0097, 0.0111] & 1.25 & 5.98 \\
                   & mAP   & 0.0203 & [0.0193, 0.0215] & 4.59 & -- \\
                   & Macro F1\textsuperscript{e} & 0.0122 & [0.0111, 0.0133] & 2.59 & -- \\
\addlinespace[2pt]
Hybrid $-$ Early   & AUROC & 0.0035 & [0.0030, 0.0039] & 0.41 & 2.12 \\
                   & mAP   & 0.0029 & [0.0023, 0.0037] & 0.64 & -- \\
                   & Macro F1\textsuperscript{e} & 0.0020 & [0.0011, 0.0029] & 0.41 & -- \\
\bottomrule
\end{tabular*}
\begin{tablenotes}[flushleft]
\scriptsize
\item Note: Headroom denotes the reduction of the remaining ranking error
$1 - \mathrm{AUROC}$ of the reference model.
\textsuperscript{e}~Exploratory; $95\%$ interval, no significance claim.
\end{tablenotes}
\end{threeparttable}
\end{table*}

\begin{table*}[!t]
\centering
\caption{Per-label DeLong tests for the three confirmatory comparisons,
computed on the one-study-per-patient subset of the test set.}
\label{tab:delong}
\small
\renewcommand{\arraystretch}{1.10}
\setlength{\tabcolsep}{6pt}
\begin{tabularx}{0.88\textwidth}{
    @{}
    >{\raggedright\arraybackslash}X
    S[table-format=2.0]
    S[table-format=3.1]
    S[table-format=2.0]
    S[table-format=3.1]
    S[table-format=1.1e2]
    @{}
}
\toprule
{\textbf{Comparison}}
& \multicolumn{2}{c}{\textbf{Uncorrected}}
& \multicolumn{2}{c}{\textbf{Bonferroni}}
& {\textbf{Median} $\boldsymbol{p}$}
\\
\cmidrule(lr){2-3}\cmidrule(lr){4-5}
& {$n$} & {\%} & {$n$} & {\%} &
\\
\midrule
Early $-$ BioViL-T & 70 & 100.0 & 69 & 98.6 & 1.7e-24 \\
Early $-$ RAD-DINO & 61 &  87.1 & 52 & 74.3 & 2.7e-11 \\
Hybrid $-$ Early   & 59 &  84.3 & 39 & 55.7 & 1.2e-3  \\
\bottomrule
\end{tabularx}
\vspace{3pt}
\begin{minipage}{0.88\textwidth}
\footnotesize
\textit{Note:}
DeLong assumes independent observations, so the test is applied to the
one-study-per-patient subset ($n = 23{,}889$ studies, $56.6\%$ of the test
set). Each comparison contributes 70 seed--label pairs (14 labels $\times$
5 seeds); $n$ denotes the number of pairs reaching significance. The
Bonferroni threshold is $0.05/14$, applied across labels within each
comparison. These tests lie outside the confirmatory family and are
reported descriptively, without significance claims.
\end{minipage}
\end{table*}

\subsection{Label-Wise Late Fusion Weights}
\label{app:late_fusion_weights}

To characterize how the three branches divide the labor across labels, we examine the label-wise weights obtained by the validation-set grid search described in Section 3.3 of the main paper. The weights show that the early-fusion branch plays the dominant role, receiving the largest mean weight across all 14 labels. However, the relative contributions of the two individual branches vary substantially among observations. RAD-DINO receives a higher weight than BioViL-T for labels such as \textit{Lung Lesion} and \textit{Pneumothorax}, whereas BioViL-T contributes more to \textit{Support Devices}, \textit{Consolidation}, and
\textit{Cardiomegaly}.

We also find that \textit{No Finding} depends almost entirely on the
early-fusion branch, with a mean weight of \(0.930\), whereas labels such as
\textit{Lung Lesion} and \textit{Support Devices} receive substantial
additional contributions from one of the individual branches. This indicates that the CONCAT representation receives the largest weight when all branches are present. However, weight magnitude does not measure indispensability. In the leave-one-branch-out analysis in Table~\ref{tab:lobo}, removing any single branch degrades both metrics with 95\% intervals excluding zero, and removing the RAD-DINO latent branch causes the largest loss of $-0.0037$ macro AUROC and $-0.0062$ mAP, exceeding the loss from removing the early-fusion branch itself. The two analyses are complementary: the weights describe how the
branches are used jointly, whereas the leave-one-out differences show that no branch is redundant within the ensemble. Weights for all 14 labels are detailed in Table~\ref{tab:late_fusion_weights}.
\begin{table*}[!t]
    \centering

     \captionsetup{
        justification=centering,
        singlelinecheck=false
    }
    
    \caption{Label-wise late-fusion weights for each label. Each value is
    reported as the mean \(\pm\) standard deviation across five seeds.}
    \label{tab:late_fusion_weights}

    \footnotesize
    \renewcommand{\arraystretch}{1.08}
    \setlength{\tabcolsep}{5pt}

    \begin{tabularx}{0.92\linewidth}{
        @{}
        >{\raggedright\arraybackslash}X
        >{\centering\arraybackslash}p{0.22\linewidth}
        >{\centering\arraybackslash}p{0.22\linewidth}
        >{\centering\arraybackslash}p{0.18\linewidth}
        @{}
    }
        \toprule
        \textbf{Label}
        & \textbf{RAD-DINO latent}
        & \textbf{BioViL-T latent}
        & \textbf{Early fusion} \\
        \midrule
        Atelectasis
            & \(0.182 \pm 0.038\)
            & \(0.168 \pm 0.038\)
            & \(0.650 \pm 0.049\) \\
        Cardiomegaly
            & \(0.196 \pm 0.059\)
            & \(0.252 \pm 0.038\)
            & \(0.552 \pm 0.063\) \\
        Consolidation
            & \(0.168 \pm 0.080\)
            & \(0.280 \pm 0.049\)
            & \(0.552 \pm 0.063\) \\
        Edema
            & \(0.224 \pm 0.059\)
            & \(0.196 \pm 0.031\)
            & \(0.580 \pm 0.049\) \\
        Cardio.
            & \(0.252 \pm 0.080\)
            & \(0.196 \pm 0.059\)
            & \(0.552 \pm 0.038\) \\
        Fracture
            & \(0.210 \pm 0.111\)
            & \(0.112 \pm 0.063\)
            & \(0.678 \pm 0.063\) \\
        Lung Lesion
            & \(0.364 \pm 0.091\)
            & \(0.056 \pm 0.031\)
            & \(0.580 \pm 0.070\) \\
        Lung Opacity
            & \(0.168 \pm 0.063\)
            & \(0.182 \pm 0.063\)
            & \(0.650 \pm 0.000\) \\
        No Finding
            & \(0.028 \pm 0.063\)
            & \(0.042 \pm 0.094\)
            & \(0.930 \pm 0.157\) \\
        Pleural Effusion
            & \(0.210 \pm 0.000\)
            & \(0.224 \pm 0.031\)
            & \(0.566 \pm 0.031\) \\
        Pleural Other
            & \(0.224 \pm 0.115\)
            & \(0.154 \pm 0.059\)
            & \(0.622 \pm 0.094\) \\
        Pneumonia
            & \(0.224 \pm 0.031\)
            & \(0.196 \pm 0.031\)
            & \(0.580 \pm 0.049\) \\
        Pneumothorax
            & \(0.266 \pm 0.059\)
            & \(0.056 \pm 0.031\)
            & \(0.678 \pm 0.038\) \\
        Support Devices
            & \(0.112 \pm 0.038\)
            & \(0.294 \pm 0.059\)
            & \(0.594 \pm 0.059\) \\
        \bottomrule
    \end{tabularx}

    \vspace{3pt}

    \begin{minipage}{0.92\linewidth}
        \footnotesize
    \end{minipage}
\end{table*}

\subsection{Representation Complementarity Analyses}
\label{app:complementarity}

All analyses in this appendix are exploratory and lie outside the
pre-specified confirmatory family; intervals are 95\% percentile
bootstrap intervals across the five seeds, and no significance claims
are made. Throughout, the RAD-DINO and BioViL-T \emph{branches} refer to
the single-source latent branches of the main configuration, evaluated
on the test set.

\paragraph{Representation- and prediction-level similarity.}
Linear centered kernel alignment (CKA) is computed between the two standardized embeddings on a fixed subsample of $n{=}20{,}000$ studies (split-independent, computed once), between the refined latents of the two branches, and between their per-label logits on the test set.. Uncertainty of the raw-embedding CKA is quantified with a study-level bootstrap ($B = 500$ resamples), and a permutation baseline is obtained by breaking the study-level pairing ($B = 500$ permutations). Linear CKA equals 1 for linearly equivalent representations; the permutation baseline yields CKA $\approx 0.002$ (mean over permutations; 95th percentile $0.002$). CKA between the raw embeddings is $0.419$ (bootstrap 95\% CI $[0.414, 0.423]$; permutation $p < 0.002$). The observed value lies far above the permutation baseline and far below 1: the two embeddings share substantial linear structure and remain far from linear equivalence. Every stage downstream of the raw embeddings is far more similar: CKA between the refined latents is 0.834 [0.828, 0.840], CKA between the branch
logits is 0.784 [0.781, 0.787], and the macro Pearson correlation of the per-study prediction errors of the two branches reaches 0.918 [0.917, 0.919]. Training on the same label set therefore drives two branches with substantially different inputs toward similar prediction functions, which bounds the additional information available to any late-fusion rule and explains why the hybrid-over-early gain in Section 5.1 of the main paper is small.

\paragraph{Per-label error structure.}
Table~\ref{tab:error_corr} reports, for each label, the Pearson and Spearman correlations of the two branches' predicted scores and the Pearson correlation of their errors, averaged over seeds. The residual disagreement is not uniformly distributed. The four labels with the lowest between-branch correlations are \textit{Fracture} (score 0.466,
error 0.817), \textit{Pleural Other} (0.614, 0.796), \textit{Pneumothorax} (0.717, 0.816), and \textit{Lung Lesion} (0.658, 0.855), all well below the corresponding macro averages of 0.787 and 0.918. Three of the four are the rare labels; the fourth, Pneumothorax, is the least prevalent mid-tier label. This concentration of complementarity on low-prevalence labels is consistent with the prevalence-stratified result in Section 5.5 of the main paper, in which the AUROC improvement of hybrid over early fusion is largest on the rare tier.

\begin{table*}[!t]
\centering
\caption{Between-branch correlation of predicted scores and errors for
the RAD-DINO and BioViL-T branches.}
\label{tab:error_corr}
\small
\renewcommand{\arraystretch}{1.10}
\setlength{\tabcolsep}{8pt}
\begin{tabularx}{0.88\textwidth}{
    @{}
    >{\raggedright\arraybackslash}X
    S[table-format=1.3]
    S[table-format=1.3]
    S[table-format=1.3]
    @{}
}
\toprule
{\textbf{Label}}
& {\textbf{Pearson (score)}}
& {\textbf{Spearman (score)}}
& {\textbf{Pearson (error)}}
\\
\midrule
Atelectasis                & 0.867 & 0.874 & 0.972 \\
Cardiomegaly               & 0.873 & 0.875 & 0.971 \\
Consolidation              & 0.835 & 0.842 & 0.948 \\
Edema                      & 0.893 & 0.891 & 0.948 \\
Enlarged Cardio.           & 0.846 & 0.837 & 0.965 \\
Fracture                   & {\bfseries 0.466} & {\bfseries 0.456} & {\bfseries 0.817} \\
Lung Lesion                & {\bfseries 0.658} & {\bfseries 0.621} & {\bfseries 0.855} \\
Lung Opacity               & 0.840 & 0.844 & 0.978 \\
No Finding                 & 0.883 & 0.856 & 0.960 \\
Pleural Effusion           & 0.876 & 0.868 & 0.937 \\
Pleural Other              & {\bfseries 0.614} & {\bfseries 0.584} & {\bfseries 0.796} \\
Pneumonia                  & 0.760 & 0.754 & 0.959 \\
Pneumothorax               & {\bfseries 0.717} & {\bfseries 0.669} & {\bfseries 0.816} \\
Support Devices            & 0.897 & 0.880 & 0.936 \\
\addlinespace[2pt]
\cmidrule{1-4}
Macro average
& \multicolumn{1}{c}{0.787}
& \multicolumn{1}{c}{0.775}
& \multicolumn{1}{c}{0.918}
\\
\bottomrule
\end{tabularx}
\vspace{3pt}
\begin{minipage}{0.88\textwidth}
\footnotesize
\textit{Note:}
Each value is the mean across five seeds. Bold marks the four labels
with the lowest between-branch correlations; lower values indicate more
residual disagreement available to fusion. The 95\% intervals of the
macro averages are [0.786, 0.789], [0.774, 0.776], and [0.917, 0.919],
respectively.
\end{minipage}
\end{table*}

\paragraph{Decision-level disagreement and oracle headroom.}
Table~\ref{tab:disagreement} cross-tabulates the thresholded decisions
of the two branches. On average the branches disagree on 11.5\% of
label decisions. An oracle that selects the correct branch whenever
exactly one branch is correct would gain 4.9 accuracy points over the
better single branch (macro either-correct accuracy minus best
single-branch accuracy), quantifying the headroom available to any
fusion rule at the decision level. The uniform fusion of the main
configuration recovers only part of this headroom, consistent with the
ablation finding in Section 5.4 of the main paper that the remaining benefit of the
fusion stage is an ensemble effect rather than an instance-wise
selection mechanism.

\begin{table*}[!t]
\centering
\caption{Decision-level agreement between the RAD-DINO and BioViL-T
branches.}
\label{tab:disagreement}
\small
\renewcommand{\arraystretch}{1.10}
\setlength{\tabcolsep}{3.5pt}
\begin{tabularx}{\textwidth}{
    @{}
    >{\raggedright\arraybackslash}X
    S[table-format=1.3]
    S[table-format=1.3]
    S[table-format=1.3]
    S[table-format=1.3]
    S[table-format=1.3]
    S[table-format=1.3]
    S[table-format=1.3]
    S[table-format=1.3]
    @{}
}
\toprule
{\textbf{Label}}
& {\textbf{Both}\,$\checkmark$}
& {\textbf{Only R}}
& {\textbf{Only B}}
& {\textbf{Both}\,$\times$}
& {\textbf{Disagr.}}
& {\textbf{Acc.\ R}}
& {\textbf{Acc.\ B}}
& {\textbf{Oracle}}
\\
\midrule
Atelectasis                & 0.658 & 0.073 & 0.065 & 0.204 & 0.138 & 0.731 & 0.723 & 0.796 \\
Cardiomegaly               & 0.655 & 0.076 & 0.077 & 0.192 & 0.153 & 0.731 & 0.732 & 0.808 \\
Consolidation              & 0.785 & 0.069 & 0.052 & 0.095 & 0.120 & 0.853 & 0.836 & 0.905 \\
Edema                      & 0.788 & 0.051 & 0.043 & 0.118 & 0.094 & 0.839 & 0.831 & 0.882 \\
Enlarged Cardio. & 0.664 & 0.101 & 0.087 & 0.148 & 0.188 & 0.765 & 0.751 & 0.852 \\
Fracture                   & 0.824 & 0.109 & 0.034 & 0.033 & 0.143 & 0.933 & 0.858 & 0.967 \\
Lung Lesion                & 0.908 & 0.042 & 0.016 & 0.035 & 0.058 & 0.950 & 0.924 & 0.965 \\
Lung Opacity               & 0.595 & 0.092 & 0.079 & 0.234 & 0.171 & 0.687 & 0.674 & 0.766 \\
No Finding                 & 0.762 & 0.053 & 0.045 & 0.141 & 0.098 & 0.815 & 0.807 & 0.859 \\
Pleural Effusion           & 0.781 & 0.052 & 0.048 & 0.119 & 0.100 & 0.833 & 0.829 & 0.881 \\
Pleural Other              & 0.925 & 0.041 & 0.014 & 0.020 & 0.055 & 0.966 & 0.939 & 0.980 \\
Pneumonia                  & 0.692 & 0.084 & 0.076 & 0.148 & 0.160 & 0.776 & 0.768 & 0.852 \\
Pneumothorax               & 0.916 & 0.033 & 0.018 & 0.033 & 0.051 & 0.949 & 0.934 & 0.967 \\
Support Devices            & 0.809 & 0.051 & 0.038 & 0.103 & 0.088 & 0.860 & 0.847 & 0.898 \\
\addlinespace[2pt]
\cmidrule{1-9}
Macro average
& \multicolumn{1}{c}{--}
& \multicolumn{1}{c}{--}
& \multicolumn{1}{c}{--}
& \multicolumn{1}{c}{--}
& \multicolumn{1}{c}{0.115}
& \multicolumn{1}{c}{--}
& \multicolumn{1}{c}{--}
& \multicolumn{1}{c}{--}
\\
\bottomrule
\end{tabularx}
\vspace{3pt}
\begin{minipage}{\textwidth}
\footnotesize
\textit{Note:}
Each value is the mean across five seeds. ``Only R'' and ``Only B''
denote studies on which exactly one branch is correct; Oracle is the
either-correct accuracy. The macro oracle headroom, computed as the
either-correct accuracy minus that of the better single branch and
averaged over labels, is 0.049. The oracle is an upper bound and is not
attainable by any fixed fusion rule.
\end{minipage}
\end{table*}

\paragraph{Leave-one-branch-out.}
To assess whether any of the three branches of the main configuration
is redundant, each branch is removed in turn and the remaining two are
fused with equal weights of $1/2$, with normalization and per-label
thresholds re-estimated on the validation set. Table~\ref{tab:lobo} reports the results. Removing any single branch degrades both macro AUROC and mAP, with all six 95\% intervals of the paired differences excluding zero; each branch therefore contributes
a unique predictive signal within the ensemble. The scope of this conclusion follows from the construction of the branches: the early-fusion branch is built from the same RAD-DINO and BioViL-T embeddings that feed the two single-source branches, so the three branches draw on two information sources. The leave-one-out differences measure the functional diversity of the three predictors, and this diversity is consistent with the ensemble-effect interpretation of the fusion stage in Section 5.4 of the main paper.

\begin{table*}[!t]
\centering
\caption{Leave-one-branch-out analysis of the main configuration.}
\label{tab:lobo}
\small
\renewcommand{\arraystretch}{1.10}
\setlength{\tabcolsep}{8pt}
\begin{tabularx}{0.88\textwidth}{
    @{}
    >{\raggedright\arraybackslash}X
    l
    c
    c
    @{}
}
\toprule
{\textbf{Configuration}}
& {\textbf{Metric}}
& {\textbf{Mean [95\% CI]}}
& {$\boldsymbol{\Delta}$ \textbf{vs.\ full [95\% CI]}}
\\
\midrule
Full (three branches)
& mAUROC & 0.8404 [0.8388, 0.8419] & \multicolumn{1}{c}{--} \\
& mAP    & 0.4666 [0.4619, 0.4713] & \multicolumn{1}{c}{--} \\
\addlinespace[2pt]
\cmidrule{1-4}
w/o RAD-DINO latent
& mAUROC & 0.8367 [0.8352, 0.8381] & $-0.0037$ [$-0.0039$, $-0.0035$] \\
& mAP    & 0.4604 [0.4558, 0.4650] & $-0.0062$ [$-0.0065$, $-0.0058$] \\
w/o BioViL-T latent
& mAUROC & 0.8382 [0.8365, 0.8399] & $-0.0022$ [$-0.0025$, $-0.0018$] \\
& mAP    & 0.4643 [0.4595, 0.4692] & $-0.0023$ [$-0.0029$, $-0.0016$] \\
w/o Early fusion
& mAUROC & 0.8377 [0.8363, 0.8390] & $-0.0027$ [$-0.0030$, $-0.0024$] \\
& mAP    & 0.4617 [0.4576, 0.4657] & $-0.0049$ [$-0.0056$, $-0.0043$] \\
\bottomrule
\end{tabularx}
\vspace{3pt}
\begin{minipage}{0.88\textwidth}
\footnotesize
\textit{Note:}
After removing a branch, the remaining two branches are fused at equal
weights of $1/2$. Values are means with 95\% intervals across five
seeds; $\Delta$ is the paired difference from the full three-branch
configuration. All six $\Delta$ intervals exclude zero. The largest
degradation on both metrics arises from removing the RAD-DINO latent
branch.
\end{minipage}
\end{table*}
\subsection{Calibration Analysis}
\label{app:calibration}
Because the fused outputs are normalized logits rather than probabilities,
we map them to probabilities with Platt scaling fitted per label on the
validation set, and evaluate calibration on the test set with the expected
calibration error (ECE) and the Brier score. The two metrics serve
different roles. ECE measures the reliability of the predicted
probabilities. The Brier score is a proper scoring rule and measures
overall probabilistic accuracy. After Platt scaling, all configurations are well calibrated, with ECE \(\leq 0.014\) across models (Table~\ref{tab:calibration}). Because every configuration satisfies the reliability requirement, we compare them on the
Brier score. Hybrid fusion attains the best Brier score at 0.0832. The margin
over the two single-source models is clear (0.0858 for RAD-DINO and 0.0887 for
BioViL-T), and their ECE is 0.0015 lower than that of hybrid fusion. Against
early fusion, the nearest configuration, the margin is 0.0004 in Brier score
with a difference of 0.0001 in ECE; the two are close on both metrics at this
resolution. We therefore state the conclusion at two levels. Hybrid fusion is
the preferred probabilistic predictor relative to either single embedding
source. Relative to early fusion, it is not worse on either metric, and the
difference is too small to support a preference on calibration grounds alone.
We report no significance test for these differences, and the calibration
analysis is exploratory.

\begin{table}[!htbp]
\centering
\caption{Calibration after per-label Platt scaling fitted on the validation
set (mean $\pm$ standard deviation across five seeds).}
\label{tab:calibration}
\begin{threeparttable}
\footnotesize
\renewcommand{\arraystretch}{1.16}
\setlength{\tabcolsep}{6pt}
\begin{tabular}{lcc}
\toprule
\textbf{Configuration} & \textbf{ECE} & \textbf{Brier score} \\
\midrule
RAD-DINO only & \(0.0125 \pm 0.0006\) & \(0.0858 \pm 0.0002\) \\
BioViL-T only & \(0.0125 \pm 0.0005\) & \(0.0887 \pm 0.0002\) \\
Early fusion  & \(0.0139 \pm 0.0000\) & \(0.0836 \pm 0.0002\) \\
Hybrid fusion & \(0.0140 \pm 0.0002\) & \(0.0832 \pm 0.0002\) \\
\bottomrule
\end{tabular}
\end{threeparttable}
\end{table}

\section{Sensitivity Analyses}
\label{app:sensitivity}

\subsection{Multi-View Aggregation}
\label{app:multiview_sensitivity}

To assess the sensitivity of the pipeline to the frontal--lateral aggregation rule, we repeat the experiments under four schemes: frontal-only and fixed frontal--lateral weights of 0.5/0.5, 0.6/0.4, and 0.7/0.3, with all other settings unchanged. Table~\ref{tab:multiview} reports macro AUROC and mAP of the main configuration under each scheme. Performance is stable across the four schemes: macro AUROC varies by at most 0.0010 and mAP by at most $0.0014$, and the 95\% intervals of all schemes overlap substantially. On macro AUROC, the three two-view schemes lie directionally slightly above frontal-only, suggesting that lateral views contribute modest additional signal, but no scheme is clearly superior to the others. The 0.6/0.4 weighting used in the main experiments is selected on the validation set and serves as a representative choice; the results indicate that the pipeline does not depend on this particular value.

\begin{table*}[!t]
\centering
\caption{Sensitivity of the main configuration to the frontal--lateral
aggregation scheme.}
\label{tab:multiview}
\small
\renewcommand{\arraystretch}{1.10}
\setlength{\tabcolsep}{8pt}
\begin{tabularx}{0.88\textwidth}{
    @{}
    >{\raggedright\arraybackslash}X
    c
    c
    c
    @{}
}
\toprule
{\textbf{Aggregation scheme}}
& {\textbf{Weights (F/L)}}
& {\textbf{mAUROC [95\% CI]}}
& {\textbf{mAP [95\% CI]}}
\\
\midrule
Frontal only
& \multicolumn{1}{c}{--}
& 0.8394 [0.8381, 0.8406] & 0.4661 [0.4615, 0.4707] \\
Frontal + lateral
& 0.5/0.5
& 0.8398 [0.8381, 0.8414] & 0.4652 [0.4605, 0.4700] \\
& 0.6/0.4
& 0.8404 [0.8388, 0.8419] & 0.4666 [0.4619, 0.4713] \\
& 0.7/0.3
& 0.8401 [0.8383, 0.8410] & 0.4663 [0.4609, 0.4707] \\
\bottomrule
\end{tabularx}
\vspace{3pt}
\begin{minipage}{0.88\textwidth}
\footnotesize
\textit{Note:} F/L denotes the fixed frontal/lateral weighting. The 0.6/0.4 scheme is the setting used in the main experiments, selected on the validation set.
\end{minipage}
\end{table*}

\subsection{Denoiser Noise Ceiling}
\label{app:sigma_sensitivity}
To assess the sensitivity of the latent-refinement block to the noise ceiling
$\sigma_{\max}$ of the training distribution
$s \sim \mathcal{U}(0, \sigma_{\max})$, we retrain the early-fusion branch
under $\sigma_{\max} \in \{0, 0.2, 0.65, 1.0, 1.25, 1.5\}$ over the same five seeds,
with all other settings unchanged. The setting $\sigma_{\max}=0$ keeps the
denoiser in place but removes the corruption, providing an architectural
control. As reported in Table~\ref{tab:sigma_sensitivity}, performance is stable across
the entire range: macro AUROC varies by at most $0.0002$ and mAP by at most
$0.0002$ across the six settings. Any ceiling in this range yields comparable
performance, and $0.65$ is adopted as a representative choice. Relative to the $\sigma_{\max} = 0$ control, all five non-zero ceilings yield
a slightly higher mAP ($0.4636 \rightarrow 0.4638$). The difference is an
order of magnitude smaller than the across-seed standard deviation and is not
statistically significant, so the choice of ceiling carries no measurable cost
or benefit within this range. These results also bound the impact of the train--inference asymmetry noted
in Section 3.3 of the main paper, in which the denoiser is
trained with $s \sim \mathcal{U}(0, 0.65)$ but applied with $s = 0$ at
inference. First, $s = 0$ lies within the support of the training noise
distribution, so the inference-time input is not out of distribution for the
denoiser. Second, as established by the ablation in Section 5.4 of the main paper, we do not attribute a
standalone gain to the denoiser: the autoencoder and the noisy-latent
objective are effective jointly, as a single latent-refinement block, and
all claims are made at the level of that block. Third, performance remains stable even when the ceiling is set to zero, so the model does not rely on the amount of corruption applied during training. This statement concerns the training objective. The map applied at inference is characterized in Appendix~\ref{app:denoiser_inference}. This analysis is exploratory and lies outside the pre-specified comparison families.

\begin{table}[H]
\centering
\caption{Sensitivity of the early-fusion branch to the noise ceiling
$\sigma_{\max}$.}
\label{tab:sigma_sensitivity}
\begin{threeparttable}
\footnotesize
\renewcommand{\arraystretch}{1.16}
\setlength{\tabcolsep}{8pt}
\begin{tabular}{ccc}
\toprule
\(\boldsymbol{\sigma_{\max}}\)
& \textbf{mAUROC}
& \textbf{mAP} \\
\midrule
0.00 & \(0.8368 \pm 0.0011\) & \(0.4636 \pm 0.0039\) \\
0.20 & \(0.8368 \pm 0.0011\) & \(0.4638 \pm 0.0039\) \\
0.65 & \(0.8369 \pm 0.0013\) & \(0.4638 \pm 0.0041\) \\
1.00 & \(0.8369 \pm 0.0013\) & \(0.4638 \pm 0.0040\) \\
1.25 & \(0.8369 \pm 0.0013\) & \(0.4638 \pm 0.0039\) \\
1.50 & \(0.8370 \pm 0.0012\) & \(0.4638 \pm 0.0039\) \\
\bottomrule
\end{tabular}
\begin{tablenotes}[flushleft]
\scriptsize
\item Note: \(\sigma_{\max}=0.65\) is the value used in the main experiments.
\end{tablenotes}
\end{threeparttable}
\end{table}

\subsection{The Latent Denoiser at Inference}
\label{app:denoiser_inference}

Appendix~\ref{app:sigma_sensitivity} shows that performance is insensitive to the
training noise ceiling. That result concerns the objective used to train the
module. It does not describe the function the module computes at inference,
where it is applied with $s = 0$. This subsection characterizes that function
directly. All analyses are exploratory, lie outside the pre-specified
comparison families, and are computed on the early-fusion branch over the same
five seeds.

\paragraph{The map at $s = 0$ is not an approximate identity.}
Table~\ref{tab:denoiser_geometry} reports the geometry of the
transformation. The mean cosine similarity between $z$ and
$\widehat{z} = g(z,0)$ is $0.851 \pm 0.009$, which corresponds to an angular
deviation of approximately $32^{\circ}$, and the relative $L_2$ deviation is
$0.547 \pm 0.015$. The two quantities are mutually consistent: under norm
preservation, an angular deviation of $32^{\circ}$ predicts a relative
deviation of $\sqrt{2 - 2\cos} = 0.546$, against an observed value of
$0.547$. The module at $s = 0$ therefore applies a substantial,
approximately norm-preserving rotation of the latent vector.

Table~\ref{tab:denoiser_geometry} also reports the reconstruction quality of
the autoencoder. The cosine similarity between $x$ and $\widehat{x}$ is
$0.770$, with a relative $L_2$ deviation of $0.643$, so the latent is a
lossy projection of the input embedding. This property is consistent with
the architecture: the classifier receives $x$ and $\widehat{z}$ as two
separate tokens (Eq.~(3) of the main paper), and the latent path is therefore not required to preserve $x$. The standard deviations of both reconstruction quantities are
below $0.001$ across seeds, so the lossy projection is a stable property of
the trained model.

\paragraph{The denoising objective is not the source of the effect.}
The geometry describes the map the module computes; the next two analyses
ask where its benefit comes from. To separate the module from the objective
that trains it, we retain the architecture of the latent denoiser, matched
by construction, and remove the stage-two denoising objective, so that the
module is trained by the classification loss alone. If the block-level gain
depended on the denoising objective, this change would remove the gain.
Instead, performance is unchanged relative to the full configuration on both
metrics ($-0.0000$ macro AUROC, $p = 0.834$; $-0.0004$ mAP,
$p = 9.53 \times 10^{-2}$; Table~\ref{tab:denoiser_inference}). The
denoising objective is therefore not the source of the module's
contribution.

\paragraph{The transformation is load-bearing at inference.}
The objective can be removed without cost; we next test whether the map
itself can be removed at inference. We train the full stack as in
configuration A3, hold the weights fixed, and route the latent two ways at
inference: through the module, giving $\widehat{z} = g(z,0)$, and around it,
giving $\widehat{z} = z$. The two routes share every trained weight, so any
difference between them is caused by applying or omitting the
transformation. Bypassing the module costs $-0.0013$ macro AUROC
($p = 6.10 \times 10^{-4}$) and $-0.0016$ mAP ($p = 1.32 \times 10^{-3}$).
Relative to the fixed threshold of $p < 0.001$, the AUROC difference is
significant and the mAP difference is not. Relative to the
Bonferroni-corrected level for the four tests in
Table~\ref{tab:denoiser_inference} ($0.05/4 = 0.0125$), both differences
are significant. The magnitude of the drop is itself interpretable: the bypassed
model reaches a macro AUROC of $0.8356$, which coincides with configuration A1 in Table 4 of the main paper, so the loss equals the full A3$\,-\,$A1 gain. Removing the transformation at inference therefore returns the model to the embedding-only baseline. The two routes place the head in different conditions. The head was fitted with $g(z,0)$ as its input, so the applied route matches its training distribution and the bypassed route departs from it. The comparison therefore measures the cost of removing the transformation from a stack trained around it, and the degradation admits two contributions: the information carried by the transformation, and the mismatch it leaves behind
at the head. The bypass establishes the transformation as part of the learned function as trained, and locates its contribution at the level of the block.

\paragraph{Reading.}
The three analyses are consistent and support a single interpretation. The
geometry shows that the module applies a large nonlinear transformation on
the latent path. The objective ablation shows that the benefit does not come
from the denoising objective. The bypass shows that the transformation
nevertheless cannot be removed at inference without losing the block-level
gain. We therefore attribute the benefit to the presence of a trained
nonlinear map between the encoder and the prediction head. This is the
block-level reading adopted in Section 5.4 of the main paper. Because the denoising objective belongs to the pre-specified design, the main configuration retains it.

\begin{table}[t]
\centering
\small
\caption{Geometry of the latent transformation at $s = 0$ and reconstruction
quality of the autoencoder, on the early-fusion branch. Values are mean
$\pm$ standard deviation across five seeds.}
\label{tab:denoiser_geometry}
\begin{tabular}{lc}
\toprule
Quantity & Mean $\pm$ SD \\
\midrule
\multicolumn{2}{l}{\textit{Latent transformation, $\widehat{z} = g(z,0)$}} \\
$\cos(z, \widehat{z})$              & 0.8510 $\pm$ 0.0093 \\
$\lVert \widehat{z} - z \rVert_2$   & 5.4684 $\pm$ 0.1642 \\
Relative $L_2$ deviation            & 0.5465 $\pm$ 0.0151 \\
$\lVert z \rVert_2$                 & 10.0288 $\pm$ 0.1116 \\
\midrule
\multicolumn{2}{l}{\textit{Autoencoder reconstruction}} \\
$\cos(x, \widehat{x})$              & 0.7701 $\pm$ 0.0004 \\
Relative $L_2$ deviation            & 0.6431 $\pm$ 0.0006 \\
\bottomrule
\end{tabular}
\vspace{0.3em}

\begin{minipage}{0.95\linewidth}
\footnotesize
Note: relative $L_2$ deviation is
$\lVert \widehat{z} - z \rVert_2 / \lVert z \rVert_2$, and analogously for
$(x, \widehat{x})$. An angular deviation of $32^{\circ}$ under norm
preservation predicts a relative deviation of $0.546$.
\end{minipage}
\end{table}

\begin{table}[H]
\centering
\caption{Inference-time role of the latent denoiser on the early-fusion
branch}
\label{tab:denoiser_inference}
\begin{threeparttable}
\footnotesize
\renewcommand{\arraystretch}{1.16}
\setlength{\tabcolsep}{5pt}
\begin{tabular}{lccc}
\toprule
\textbf{Configuration} & \textbf{Mean} & \(\boldsymbol{\Delta}\) & \(\boldsymbol{p}\) \\
\midrule
\multicolumn{4}{l}{\textit{Macro AUROC}} \\
Applied, \(\widehat{z} = g(z,0)\) & 0.8369 & -- & -- \\
Bypassed, \(\widehat{z} = z\)     & 0.8356 & \(-0.0013\) & \(6.10 \times 10^{-4}\) \\
Objective removed                 & 0.8369 & \(-0.0000\) & \(8.34 \times 10^{-1}\) \\
\midrule
\multicolumn{4}{l}{\textit{mAP}} \\
Applied, \(\widehat{z} = g(z,0)\) & 0.4638 & -- & -- \\
Bypassed, \(\widehat{z} = z\)     & 0.4620 & \(-0.0016\) & \(1.32 \times 10^{-3}\) \\
Objective removed                 & 0.4633 & \(-0.0004\) & \(9.53 \times 10^{-2}\) \\
\bottomrule
\end{tabular}
\begin{tablenotes}[flushleft]
\scriptsize
\item Note:  $\Delta$ is the paired difference from the full configuration across
five seeds; $p$ is a two-sided paired $t$-test.
\end{tablenotes}
\end{threeparttable}
\end{table}

\section{Supplementary Ablation Analyses}
\label{app:supp_ablation}

\begin{table*}[!t]
\centering
\small
\caption{Two-factor decomposition of the A8~\(-\)~A9 contrast: representation
quality crossed with the late-fusion rule. Cell values are mean \(\pm\)
standard deviation across five seeds.}
\label{tab:fusion_factorial}
\begin{tabular}{llcc}
\toprule
Representation & Fusion rule & macro AUROC & mAP \\
\midrule
Refined & Uniform (\(1/3\))  & \(0.8404 \pm 0.0013\) & \(0.4666 \pm 0.0038\) \\
Refined & Learned label-wise & \(0.8398 \pm 0.0011\) & \(0.4674 \pm 0.0039\) \\
Raw     & Uniform (\(1/3\))  & \(0.8362 \pm 0.0012\) & \(0.4570 \pm 0.0036\) \\
Raw     & Learned label-wise & \(0.8364 \pm 0.0010\) & \(0.4621 \pm 0.0033\) \\
\midrule
\multicolumn{2}{l}{\textit{Effect}} & \(\Delta\) AUROC (\(p\)) & \(\Delta\) mAP (\(p\)) \\
\midrule
\multicolumn{2}{l}{Representation \(\mid\) uniform} & \(+0.0042\) (\(1.73\times10^{-5}\))\,\checkmark & \(+0.0096\) (\(1.46\times10^{-6}\))\,\checkmark \\
\multicolumn{2}{l}{Representation \(\mid\) learned} & \(+0.0034\) (\(9.65\times10^{-5}\))\,\checkmark & \(+0.0053\) (\(3.65\times10^{-5}\))\,\checkmark \\
\multicolumn{2}{l}{Fusion rule \(\mid\) refined}    & \(-0.0006\) (\(2.50\times10^{-2}\))\,\(\times\) & \(+0.0008\) (\(3.38\times10^{-3}\))\,\(\times\) \\
\multicolumn{2}{l}{Fusion rule \(\mid\) raw}        & \(+0.0002\) (\(2.73\times10^{-1}\))\,\(\times\) & \(+0.0051\) (\(1.53\times10^{-5}\))\,\checkmark \\
\multicolumn{2}{l}{Interaction}                     & \(-0.0008\) (\(2.82\times10^{-2}\))\,\(\times\) & \(-0.0043\) (\(1.36\times10^{-4}\))\,\checkmark \\
\bottomrule
\end{tabular}

\vspace{0.4em}
\begin{minipage}{0.95\textwidth}
\footnotesize
Note: Representation effects are refined \(-\) raw; fusion-rule effects are
learned \(-\) uniform. \(p\)-values are two-sided paired \(t\)-tests across
seeds. A \checkmark denotes significance at the fixed threshold \(p < 0.001\),
which is stricter than the Bonferroni-corrected level for the ten tests in this
table.
\end{minipage}
\end{table*}

This appendix reports two secondary analyses of the fusion stage of
Section 5.4 of the main paper. Both were carried out after the
eleven-configuration ablation and neither belongs to the pre-specified
confirmatory family. Difference tests use the fixed threshold \(p < 0.001\)
applied throughout the paper. The equivalence test uses the separate criterion
declared in Section 3.4 of the main paper.
\subsection{Factorial Decomposition of the A8\,--\,A9 Contrast}
\label{app:fusion_factorial}

The A8~\(-\)~A9 contrast in Section 5.4 of the main paper varies the branch
representation and the fusion rule simultaneously. This subsection separates
the two factors. The decomposition is a secondary analysis, carried out after
the eleven-configuration ablation, for the sole purpose of resolving that
confound. All four cells share the data partition, training schedule,
evaluation protocol, and five seeds of the main experiments.

Table~\ref{tab:fusion_factorial} reports the cell means and the simple effects.
The refined\,+\,uniform cell is configuration A8 and the raw\,+\,learned cell
is configuration A9, so the reported contrast equals the representation effect
at uniform weights minus the fusion-rule effect at raw representations
(\(+0.0042 - 0.0002 = +0.0040\)), which recovers the value in
Table 4 of the main paper. The raw + uniform cell is configuration (iv) of Section 3.1 of the main paper, that is, hybrid fusion without latent refinement. The interaction is defined as the fusion-rule effect at the refined level minus that effect at the raw level. Representation quality is the dominant factor. Refined branches outperform raw
branches under both fusion rules and on both metrics, and all four simple
effects reach the declared threshold (\(p \leq 9.65\times10^{-5}\)).

The fusion rule matters only when the representations are raw. At the raw
level, learned weights improve mAP over uniform weights by \(+0.0051\)
(\(p = 1.53\times10^{-5}\)). At the refined level, the same substitution yields
\(+0.0008\) mAP (\(p = 3.38\times10^{-3}\)) and \(-0.0006\) AUROC
(\(p = 2.50\times10^{-2}\)), neither of which reaches the threshold.

The interaction is significant on mAP (\(-0.0043\),
\(p = 1.36\times10^{-4}\)) and carries the same sign on AUROC without reaching
the threshold (\(-0.0008\), \(p = 2.82\times10^{-2}\)). A five-seed design has
limited power for an interaction test. We therefore report estimates and signs
alongside the \(p\)-values and do not treat the AUROC result as evidence of
absence. The reading supported by both metrics is that learned label-wise weights compensate for unrefined branches, and their benefit falls by an order of magnitude once latent refinement is applied. The two mechanisms address the same deficiency, and only
latent refinement is retained in the final configuration.

\subsection{Equivalence of the Deep Ensemble and Uniform Fusion}
\label{app:fusion_tost}

Configuration A11 averages the logits of three independently initialized
early-fusion models. Configuration A8 fuses the three refined branches with
uniform weights. The paired difference between them is not significant on
either metric, which by itself supports no conclusion. We therefore applied two
one-sided tests against the margins of \(\pm0.0035\) macro AUROC and
\(\pm0.0029\) mAP declared in Section 3.4 of the main paper, at
\(\alpha = 0.05\) with \(90\%\) confidence intervals.

On macro AUROC the observed difference is \(-0.0005\), with 90\% CI
\([-0.0010, +0.0001]\) and \(p_{\mathrm{TOST}} = 1.59\times10^{-4}\). The
interval is contained within the margin and includes zero.

On mAP the observed difference is \(+0.0008\), with 90\% CI
\([+0.0001, +0.0014]\) and \(p_{\mathrm{TOST}} = 1.11\times10^{-3}\). The
interval is contained within the margin and excludes zero. The deep ensemble is
therefore slightly better than uniform fusion on mAP, by an amount smaller than
the smallest effect this study treats as meaningful. We describe the two
configurations as equivalent at the declared margin, not as identical.

Equivalence at this margin does not imply that the fusion of heterogeneous
branches and the ensembling of homogeneous models are interchangeable in
general. It supports the narrower claim made in
Section 5.4 of the main paper: within the resolution of the present study, the
fusion stage contributes no more than plain ensembling contributes.

{
    \small
    \bibliographystyle{ieeenat_fullname}
    \bibliography{references}
}